\documentclass{pkumelon}
\usepackage{lmodern}

\DeclareFontShape{T1}{bytesans}{b}{n}{<-> s * [1] pkumelon/bytesans}{}
\DeclareFontShape{T1}{bytesans}{bx}{n}{<-> s * [1] pkumelon/bytesans}{}
\setcitestyle{authoryear,round,citesep={;},aysep={,},yysep={;}}
\renewcommand{\titlefont}{\fontsize{17}{20}\selectfont\raggedright}
\renewcommand\affiliationformat[2][]{{\affiliationfont {\sffamily $^{#1}$#2}}}
\makeatletter
\renewcommand\authorformat[2][]{%
  \authorfont {\sffamily\mbox{\ifstrempty{#1}{#2}{#2$^{#1}$}}}%
}
\makeatother
\fancypagestyle{firststyle}{%
  \fancyhf{}
  \renewcommand{\headrulewidth}{0pt}
  \fancyhead[L]{\includegraphics[width=18mm]{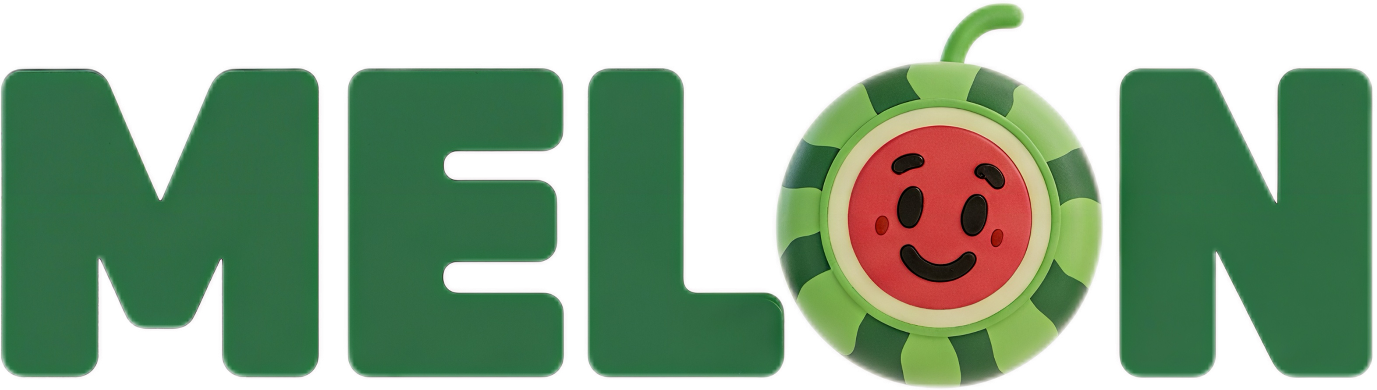}}
  \fancyhead[R]{\sffamily\textcolor[HTML]{4D4D4D}{Preprint}}
}

\usepackage{amsmath,amssymb,amsthm}
\usepackage{booktabs}
\usepackage{multirow}
\usepackage{longtable}
\usepackage{needspace}
\usepackage{array}
\usepackage{ragged2e}
\usepackage{algorithm}
\usepackage{algpseudocode}
\usepackage{graphicx}
\usepackage{xcolor}
\usepackage{colortbl}
\usepackage{tikz}
\usetikzlibrary{arrows.meta,positioning,fit,backgrounds,calc}
\definecolor{acc}{HTML}{2E7D5B}
\definecolor{rej}{HTML}{B5532A}
\definecolor{linkblue}{HTML}{2F5DA8}
\definecolor{ink}{HTML}{1A1A1A}
\definecolor{rulegray}{HTML}{8A8A8A}
\definecolor{panelbg}{HTML}{F4F4F2}
\usepackage[bottom]{footmisc}
\usepackage{hyperref}
\usepackage{xurl}  
\usepackage{fvextra} 
\usepackage{placeins} 

\usepackage{tabularx}
\newcolumntype{R}{>{\raggedleft\arraybackslash}X}
\colorlet{tablehighlight}{reaslab!12!white}
\newcommand{\tablehead}[1]{\textcolor{reaslab}{\textbf{#1}}}
\newcommand{\MelonTableSetup}{%
  \arrayrulecolor{reaslab!75!white}%
  \setlength{\heavyrulewidth}{0.8pt}%
  \setlength{\lightrulewidth}{0.35pt}%
  \setlength{\cmidrulewidth}{0.3pt}%
  \setlength{\extrarowheight}{1pt}%
}
\AtBeginEnvironment{tabular}{\MelonTableSetup}
\AtBeginEnvironment{tabular*}{\MelonTableSetup}
\AtBeginEnvironment{longtable}{\MelonTableSetup}
\DeclareCaptionFormat{melontable}{%
  {\sffamily\textcolor{reaslab}{\textbf{#1}}}\enspace #3}
\hypersetup{
  pdftitle={ProofLoom: Proof-Obligation-Driven Theory Construction for Autoformalizing Research-Level Stochastic Optimization},
  pdfauthor={Feiming Wang, Daibo Li, Kun Yuan},
  pdfkeywords={stochastic optimization, formalization, Lean, proof obligations, machine-checked proofs, formal methods},
  colorlinks=true,
  linkcolor=linkblue,
  citecolor=acc,
  urlcolor=linkblue
}

\theoremstyle{definition}

\theoremstyle{plain}

\newcommand{\lean}[1]{\texttt{#1}}
\newcommand{\indexlabel}[1]{\textcolor{acc}{\textbf{#1}}}

\newcommand{\NAlgos}{33}
\newcommand{\NBenchAlgos}{15}

\newcommand{\NDecls}{2{,}399}
\newcommand{\NLemmas}{1{,}983}
\newcommand{\NLeanLOC}{490{,}693}
\newcommand{\NPublicAlgos}{32}
\newcommand{\NPublicLeanLOC}{414{,}628}
\newcommand{\NPrivateLeanLOC}{76{,}065}
\newcommand{\NSOptLOC}{109{,}634}
\newcommand{\NLibCatalog}{2{,}294}
\newcommand{\NJointAssessments}{180}
\newcommand{\NModelRatings}{1,260}
\newcommand{\NHumanRatings}{105}

\newcommand{\NSourceIssues}{28}
\newcommand{\NSourceOriginalIssues}{25}
\newcommand{\NSourceRevisedIssues}{3}
\newcommand{\NSourceIssueAlgorithms}{22}
\newcommand{\NSourceAuditAlgorithms}{33}

\title{ProofLoom: Proof-Obligation-Driven Theory Construction for Autoformalizing Research-Level Stochastic Optimization}

\author[1,*]{Feiming Wang}
\author[2]{Daibo Li}
\author[2,\dagger]{Kun Yuan}
\affiliation[1]{Nankai University}
\affiliation[2]{Peking University}

\abstract{
Formalizing research-level stochastic optimization in Lean requires a Lean model of the algorithm and domain theory that connects foundational libraries to its convergence proof. Published proofs often compress these connections, and whether a Lean model supports the complete proof may become clear only as the proof proceeds. Revising the model to restore provability, however, can change the mathematical claim. We introduce \textbf{ProofLoom}, a fully automated LLM-agent system for Proof-Obligation-Driven Theory Construction. Given a published algorithm, its target theorem, and its source proof, ProofLoom autonomously constructs the Lean model and the mathematical infrastructure required to complete the proof. Open proof obligations drive the joint development of definitions, interfaces, supporting lemmas, and proof plans. Signature contracts record the published evidence and derived obligations for each revision of the Lean model, and an independent Judge rejects revisions that add unsupported assumptions or weaken the theorem. Planner expands the published argument into intermediate claims, and Audit checks whether the Lean proof follows that argument. Across tasks, ProofLoom accumulates both verified mathematics and construction experience in \textbf{SOptLib}. It extracts and generalizes reusable results from certified developments, verifies their integration, and records modeling decisions and failed proof routes. Subsequent tasks retrieve these results and records to guide modeling and proof construction, extending the knowledge available to later formalizations. On fifteen textbook and research-paper tasks, ProofLoom obtains mean human ratings of 6.3/7 and 6.4/7, compared with 4.9/7 and 5.0/7 for the strongest of six baselines. Across \NAlgos{} developments, it produces \NLeanLOC{} lines of algorithm-local Lean code with no \texttt{sorry}. The formalizations also expose \NSourceIssues{} incorrect formulas, proof gaps, and algorithm--analysis mismatches in published sources across \NSourceIssueAlgorithms{} developments, each with checked evidence.
Code and supplementary materials are available at
\url{https://github.com/Trace231/ProofLoom}.
}

\begin{document}
\maketitle
\begingroup
\renewcommand{\thefootnote}{\fnsymbol{footnote}}
\footnotetext[1]{Work done while interning at Peking University}
\footnotetext[2]{Corresponding author: \email{kunyuan@pku.edu.cn}}
\endgroup


\suppressfloats[t]
\begin{figure}[t]
  \centering
  \includegraphics[width=0.938741\textwidth,trim=12bp 0bp 4bp 24bp,clip]{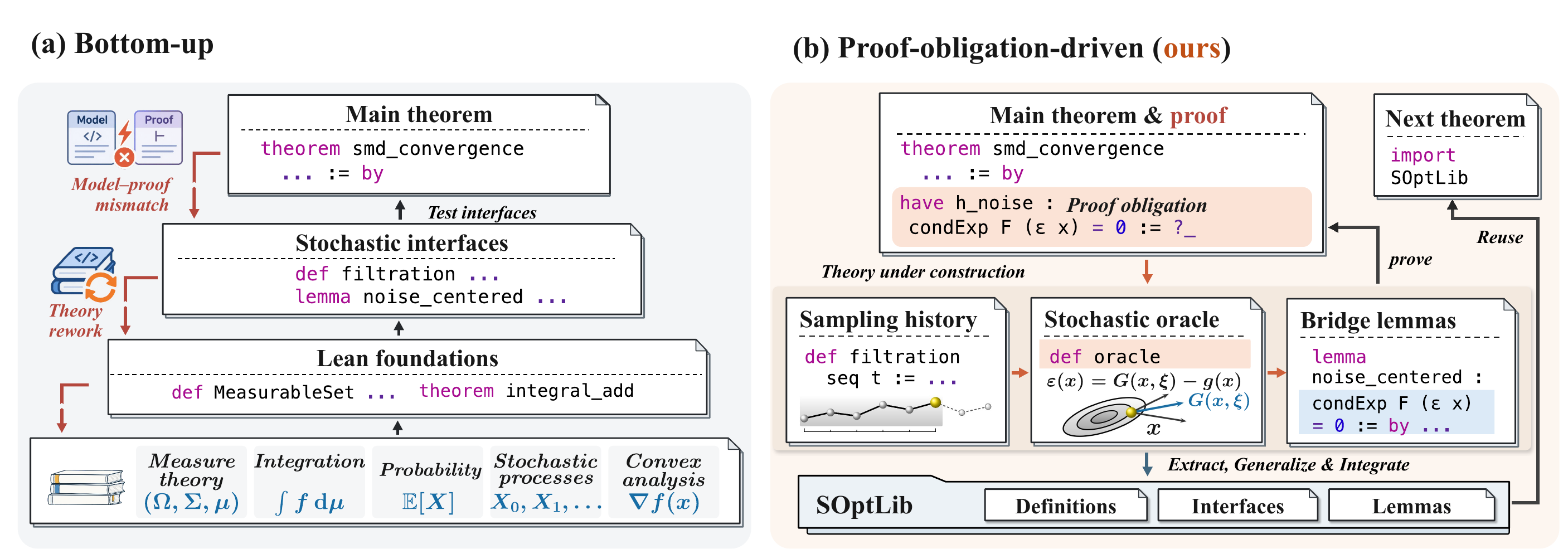}
  \caption{\textbf{Two construction orders.}
  Bottom-up methods build theory first; ProofLoom constructs definitions,
  interfaces, and lemmas as required by the target proof.}
  \label{fig:construction-order}
\end{figure}

\section{Introduction}
\label{sec:intro}

Stochastic optimization underlies the training of modern machine learning
models, from stochastic gradient methods to variance-reduced and adaptive
algorithms. The convergence guarantees of these algorithms are established in
published proofs and checked mainly through peer review. A machine-checked Lean
formalization provides a stronger guarantee. Producing one involves more than
translating the theorem statement into Lean. It requires a \emph{Lean model} of
the algorithm, which specifies its updates, randomness, assumptions, and target
theorem. It also requires the domain theory used in the convergence analysis.
Mathlib provides general foundations in probability and convex analysis
\citep{mathlib}, but the definitions, interfaces, and bridge lemmas that
connect these foundations to stochastic optimization are largely absent.
We study how LLM agents can produce such a formalization, including its Lean
model and the missing domain theory, from a published research-level
algorithm, its convergence theorem, and its proof.

Formalizing a research-level algorithm together with its convergence analysis
raises two difficulties. First, published proofs often compress the connections among measure-theoretic probability, convex and real analysis, and the algorithm into a few steps, which the formalization
must make explicit through interfaces and bridge lemmas. Second, the same algorithm
admits several Lean models, and whether a Lean model supports the complete
proof may become clear only when the proof reaches a step that depends on it.
Stochastic mirror descent illustrates both difficulties \citep{lan2020foml}. A
key step of its analysis shows that a noise term at the current iterate has
zero expectation. The oracle assumption states
$\mathbb{E}_{\xi}[G(x,\xi)] = g(x)$ for each fixed $x$, whereas the proof uses
$\mathbb{E}[G(x_t,\xi_t)\mid\mathcal{F}_{t-1}] = g(x_t)$ at the random iterate
$x_t$, where $\mathcal{F}_{t-1}$ is generated by the past samples
$\xi_1,\ldots,\xi_{t-1}$. The published proof takes this step directly, but the formalization needs a bridge lemma, with measurability and integrability conditions, to derive the identity from the fixed-query assumption. The lemma is provable only if the Lean model specifies how $x_t$ depends on past samples and how $\xi_t$ is drawn; a Lean model omitting them cannot support the step, a gap exposed only when the proof reaches it.

These difficulties make the order of construction in Lean consequential, as each order carries a distinct risk. Bottom-up development builds the supporting theory along its dependencies before attempting the target theorem and proof \citep{immler2018thesis,ericson2026scott}. Because the Lean model is fixed at the outset, a mismatch between the model and the proof may surface only after substantial development, requiring the theory built on the model to be reworked. Top-down development starts from the target theorem and exposes such mismatches early, but must revise the Lean model as the proof reveals missing relations, such as the sampling dependencies above. Revisions made to restore provability can change the mathematical claim, for example by assuming a fact that the published proof derives or by weakening its conclusion \citep{dollmsgame2026}. The challenge is to let the target proof drive theory construction and Lean model revision while preserving the published result.

We introduce \textbf{ProofLoom}, a fully automated LLM-agent system for Proof-Obligation-Driven Theory Construction in stochastic optimization. Given a published algorithm, its target theorem, and its source proof, ProofLoom autonomously constructs the Lean algorithm model and develops the mathematical infrastructure required to complete the proof. Open obligations in the target proof determine which definitions, interfaces, and lemmas to reuse or construct next, driving the joint development of the model, supporting theory, and proof plans. When proof construction stalls, ProofLoom diagnoses the cause using the source proof, the current Lean state, and prior proof attempts, then autonomously selects and executes the next steps: proving a missing fact, revising the proof route, or reconstructing the Lean model. Lean checks every construction. Two mechanisms keep this process faithful to the published result. Signature contracts govern revisions of the Lean model: each contract records the changed declarations, the supporting published passage, and the derived claims the change introduces. Derived claims remain proof obligations rather than assumptions, and an independent Judge rejects revisions that add unsupported assumptions or weaken the theorem. Planner--Audit governs proof construction: Planner expands the published argument into intermediate claims, and Audit traces the realized Lean dependencies against that argument to locate missing mathematical links and unnecessary proof branches.

Across algorithm formalizations, ProofLoom continually accumulates reusable mathematical knowledge and the experience of constructing it through \textbf{SOptLib}. After a development is certified, its reusable definitions and lemmas are generalized across algorithms, proved in Lean, and rechecked in the original development. SOptLib also contains a natural-language knowledge base that records modeling decisions, failed proof routes, and supporting evidence. Subsequent formalizations retrieve these mathematical results and construction records to guide modeling and proof planning. Once certified, these developments contribute newly constructed reusable results and experience back to SOptLib, forming a cycle of construction, accumulation, and reuse across algorithms.

We evaluate ProofLoom against six baselines on fifteen tasks: ten algorithms
from Lan's monograph (FOML) and five from research papers. ProofLoom obtains
mean human ratings of 6.3/7 on FOML and 6.4/7 on research-paper tasks, compared
with 4.9/7 and 5.0/7 for the strongest baseline. It also achieves the highest
mean score in all 16 combinations of task group, evaluation protocol, and model
judge. In an ablation on 43 recorded obstructions, the full system correctly
diagnoses the obstruction and proposes a faithful repair in 33 cases, compared
with 29 for either ablation; removing Judge raises incorrect repairs from one
to six. Across \NAlgos{} developments, ProofLoom produces \NLeanLOC{} lines of
algorithm-local Lean code with no \texttt{sorry}, and SOptLib contains
\NSOptLOC{} lines and \NDecls{} declarations. The formalizations also expose
\NSourceIssues{} discrepancies in published sources across
\NSourceIssueAlgorithms{} developments, including a counterexample to a
textbook corollary.

Our contributions are as follows.

\begin{enumerate}
  \item \textbf{Automated proof-obligation-driven theory construction.}
  We introduce a fully automated system that jointly constructs Lean
  algorithm models and supporting theory from source algorithms,
  theorem statements, and proofs, guided by the obligations of the
  target proof.

  \item \textbf{Construction-time faithfulness control.}
  Signature contracts, Judge review, and Planner--Audit govern revisions
  of the Lean model against the published result.

  \item \textbf{Continual knowledge accumulation through SOptLib.}
  The system extracts, generalizes, and integrates verified mathematics
  and construction experience from certified developments. Subsequent
  tasks retrieve and reuse this accumulated knowledge to guide modeling
  and proof construction.

  \item \textbf{Verified artifacts and source findings.}
  We develop Lean formalizations of \NAlgos{} algorithms and catalogue
  \NSourceIssues{} discrepancies in published sources with checked evidence.
\end{enumerate}

\section{Proof-Obligation-Driven Theory Construction}
\label{sec:topdown}

Given a source algorithm $A$, its main convergence theorem $T$, and proof $P$,
ProofLoom formalizes the complete algorithm from $S=(A,T,P)$. The development includes the algorithm
model, the target proof, and the domain infrastructure required by the proof. Throughout
construction, its objects, assumptions, and conclusion must retain their correspondence
to the source. The current Lean statement $T_t^\star$ may be revised under a recorded
blocker and signature-contract review.

\subsection{Constructing Theory from the Target Proof}
\label{sec:topdown:order}

Top-down construction starts from the algorithm's main convergence theorem and identifies
the infrastructure required by its proof (Figure~\ref{fig:construction-order}).
ProofLoom introduces the algorithm model and theorem while supporting theory is
incomplete. Current Lean obligations and recorded proof attempts determine what to build next. ProofLoom expands obligations along the
source proof and reuses Mathlib and SOptLib where applicable. Remaining
obligations expose a \emph{dependency frontier}: the missing definitions, interfaces, and
bridge lemmas needed by the proof. Proof attempts can reveal missing lemmas or interfaces
that cannot express a step in the source proof. This guides proof work and proposed model revisions, with infrastructure built
and checked in dependency order.

In stochastic mirror descent, the main convergence proof requires a one-step descent
bound. That bound uses a conditional expectation identity at the random iterate.
Expressing this step requires a filtration for the sampling history, adapted iterates,
and the relevant measurability and sampling relations. With these relations expressed, the next task is proving the bridge lemma.
If the oracle interface cannot express a source dependency,
the mismatch supports reconstruction.

\begin{figure}[!t]
  \centering
  \includegraphics[width=0.98\linewidth,trim=9bp 5bp 11bp 51bp,clip]{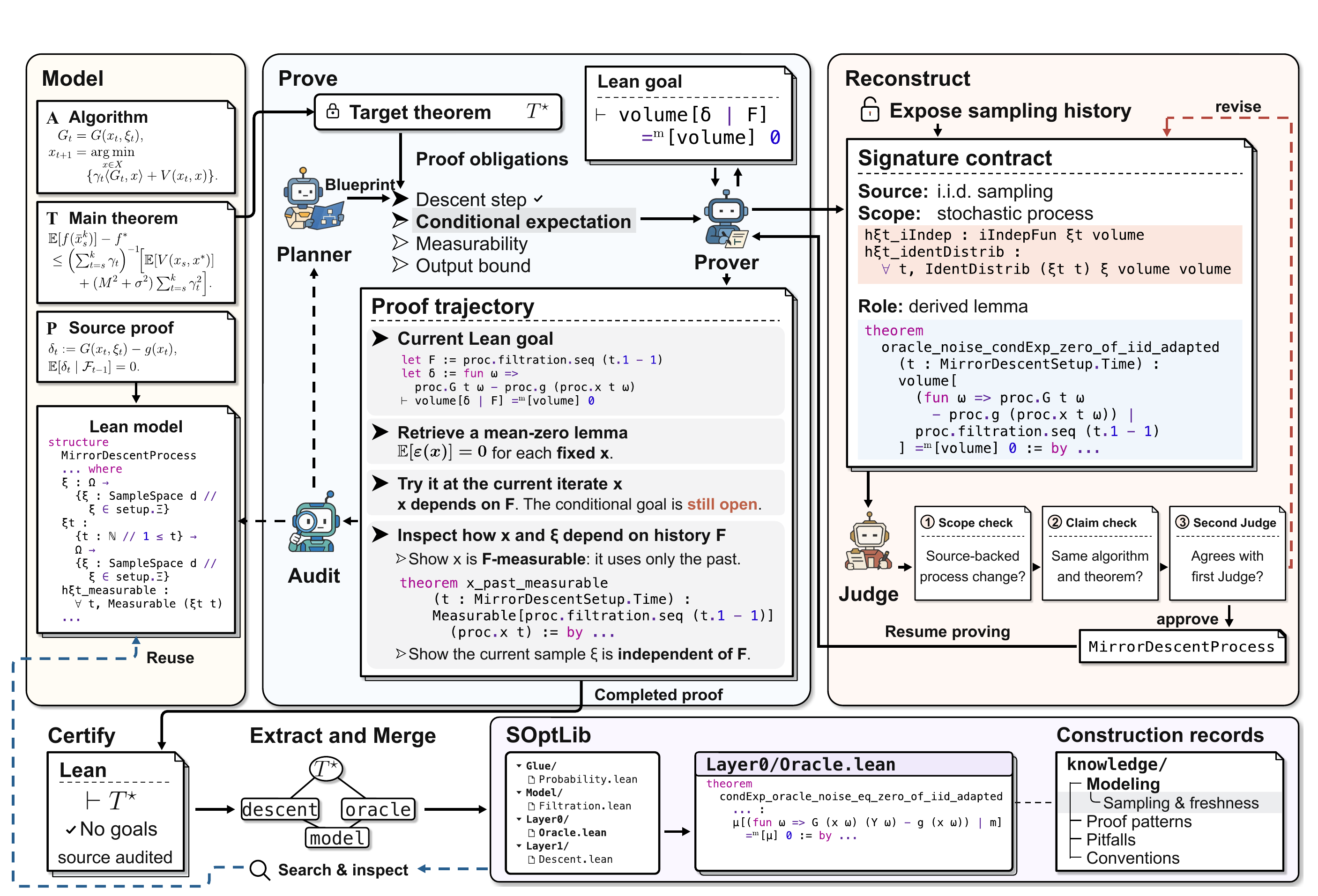}
  \caption{\textbf{ProofLoom system overview.}
  Model defines the algorithm and theorem; Construct builds proofs and reviews
  revisions; Learn certifies developments and updates SOptLib.}
  \label{fig:pipeline}
\end{figure}

\subsection{ProofLoom Pipeline}
\label{sec:topdown:stages}

\noindent\textbf{Model.}
Model uses the source and retrieved SOptLib entries to define canonical objects,
primitive assumptions, algorithm state and updates, stochastic semantics, the output
rule, and theorem statement. Source assumptions enter the setup; derived properties
remain proof obligations. The resulting declarations establish the protected interface.

\noindent\textbf{Construct.}
The read-only Planner uses Lean evidence, the source proof, retrieved results, and prior
attempts to develop a blueprint for the current proof gap or report a blocker.
Prover follows the blueprint to build proofs and local helpers under the protected
interface. Audit traces the Lean dependencies and checks correspondence with the
source proof step by step. Proof attempts and review outcomes are retained
for subsequent planning and audit.
A blocker records the source dependency or audited route issue requiring
reconstruction, together with prior repair evidence. Refactor develops a candidate
revision and its signature contract. Judge reviews the source correspondence and
affected dependencies, using Lean evidence to guide further revisions
(Section~\ref{sec:faithful}). The resulting candidate, review, and remaining
obligations inform subsequent proof work.

\noindent\textbf{Learn.}
After all Lean obligations close, Certify checks the development's canonical
entry point, full dependencies, source correspondence, and audit evidence.
Upon certification, Extract and Merge add reusable mathematics and reviewed
construction experience to SOptLib.
Algorithm~\ref{alg:full-pipeline} in Appendix~\ref{app:control-loop} gives the
complete control loop.

\label{sec:topdown:reuse}

Agents retrieve SOptLib definitions, lemmas, and construction
records to guide modeling and proof planning. Reuse requires instantiating results
and proving their hypotheses. ProofLoom constructs missing support, generalizes reusable
results, checks them in Lean, and verifies their use in the original development.
These results and reviewed records enter SOptLib for later algorithms
(Section~\ref{sec:assets}).

\section{Governing Proof--Model Co-evolution}
\label{sec:faithful}

Proof-obligation-driven construction must identify which mathematics the target proof
needs and how to express it faithfully in Lean. ProofLoom addresses these questions
through two mechanisms. Signature contracts make model revisions explicit and subject
them to adversarial review. Planner--Audit combines forward planning and backward
auditing in a dual mechanism. It compares the source argument with actual proof
dependencies to locate missing infrastructure and revise failed routes.
Both record the obligation, source passage, and Lean evidence supporting each decision.

\subsection{Signature Contracts for Model Revision}
\label{sec:faithful:contracts}

The development must preserve the source algorithm, its assumptions, and its convergence
claim. Its current Lean statement $T_t^\star$ and supporting interfaces can evolve as
the proof exposes missing structure, while retaining this source correspondence.
Ordinary proof work preserves the protected signatures and adds proofs and internal
helpers. Planner reports a blocker
when recorded evidence calls for interface reconstruction or an audited revision of
the dependency structure. The report identifies the affected obligation, its source
context, and the proof or infrastructure repairs already attempted. Repeated failure
prompts investigation; reconstruction depends on the diagnosed cause.

\noindent\textbf{Faithfulness as a contract.}
The signature contract records the affected declarations, proposed changes,
mathematical roles, source evidence, and dependent proofs that require checking.
Refactor develops a candidate revision using this record, the failed proof attempts,
and Judge's feedback. The contract records the reason for each change throughout
reconstruction (Appendix~\ref{app:signature-contract-workflow}).

\noindent\textbf{Adversarial review.}
We develop Judge's checklist from established principles in the formalization literature
\citep{barendregt2005challenge,wiedijk2012pollack}. Judge uses these principles to check
source correspondence, assumptions, object construction, and proof dependencies.
For each added premise, Judge reads the cited passage and checks whether the source
states it as an assumption or derives it in the proof. Derived facts retain proof
obligations. For a revised object, Judge checks its definition and proved properties.
This includes any bridge needed to connect a new representation to the source object. It also checks whether reuse of Mathlib or SOptLib preserves the
source assumptions and output rule. A library interface may require a new bridge.

\noindent\textbf{Iterative reconstruction.}
Judge reviews candidates against the source and contract.
Its feedback identifies the affected declaration, source passage,
and required construction or proof. Refactor uses this feedback to revise
the candidate. Review covers accumulated changes and unresolved issues.
Judge checks signature changes against the recorded scope and examines definition bodies.

The contract $C$ records the derived claims required by the revision as $\mathcal D_C$.
For the revised development $L'$, let $\mathcal P(L')$ contain claims with completed
Lean proofs and checked dependencies, and let $\mathcal O(L')$ contain explicitly
recorded proof obligations. Let $\mathcal A_{\mathrm{new}}(L')$ contain additional
assumptions introduced in source-facing interfaces. Reconstruction must preserve
\[
\mathcal D_C\subseteq\mathcal P(L')\cup\mathcal O(L'),
\qquad
\mathcal D_C\cap\mathcal A_{\mathrm{new}}(L')=\varnothing .
\]
Judge checks each claim's role in the source and follows its use through the
revised declarations. Lean checks the candidate, while dependency inspection
identifies unfinished proofs. An approved revision resolves the reported interface
or proof-route issue and passes the source-correspondence and Lean checks.
Remaining proof obligations return to Planner.
Appendix~\ref{app:control-loop} gives the full loop.

For example, a source proof may use the optimality condition of a proximal update
defined by minimization. Judge traces this condition to the update definition and
identifies the required derivation. Refactor constructs the update interface and
introduces the corresponding lemma. Any unfinished proof remains in $\mathcal O(L')$.
Planner schedules the remaining work, and Audit checks how the convergence proof
uses the lemma. After the lemma is proved, later targets can reuse it and the interface.

If investigation establishes a gap in the source claim, the record retains the
original statement and supporting evidence. Any corrected result is reported
separately. See Appendix~\ref{app:faithfulness-audit} for details.

\subsection{Planner--Audit for Proof Construction}
\label{sec:faithful:planner-audit}

Planner develops a forward blueprint from the source proof. For each gap,
it identifies intermediate claims, their underlying definitions and assumptions,
and intended dependencies. Retrieved results and prior attempts guide this
decomposition. Prover implements the blueprint through proofs and local helpers.
The trajectory records constructions, Lean goals, and outcomes, including failures.

\noindent\textbf{Tracing the realized argument.}
Audit works backward from the Lean declarations and execution evidence to the source
steps they are meant to establish. It checks why each unresolved helper is needed
by the target proof. This check uses the source argument, actual Lean dependencies,
and evidence from attempted library applications. It also checks whether the helpers'
assumptions and conclusions support the corresponding source steps. A source step
can require several Lean lemmas; the audit examines why that expansion is needed.
For each result invoked by the source, Audit searches for an applicable interface
and checks its actual use. If the interface's proof depends on the unresolved helper,
using it leaves that helper to be proved.

Audit also checks whether another applicable source-level result
can resolve the target without the current helper chain. A locally difficult lemma
may come from an unnecessarily strong decomposition or an obsolete proof branch.
Audit requests evidence to continue a branch, including failed interface
applications, incompatible hypotheses, or a missing mathematical dependency.
These checks distinguish infrastructure required by the source argument from
extra work introduced by the plan.

\noindent\textbf{Diagnosis, discovery, and correction.}
The audit identifies the source step, Lean obligation, and evidence supporting its
diagnosis. Planner uses it to revise the blueprint, reuse an existing result, or
concentrate construction on a missing bridge. When the remedy requires interface
or dependency reconstruction, Planner reports a blocker for Refactor--Judge review.
The construction history retains failed routes and their diagnoses, so later attempts
can address the recorded cause. Together, forward planning and backward audit enable
the system to uncover implicit proof steps, investigate missing conditions, and
test proposed repairs through Lean proofs or counterexamples.

\subsection{Worked Example: SAM's Gaussian KL Bridge}
\label{sec:sam-construction}

SAM's PAC-Bayes argument requires a Gaussian KL calculation. For posterior
$Q=\mathcal N(w,\sigma^2I_d)$ and prior $P_j=\mathcal N(0,v_jI_d)$, with
$\sigma>0$ and $v_j>0$, the target identity is
\[
\operatorname{KL}(Q\Vert P_j)
=\frac12\left(
\frac{d\sigma^2+\lVert w\rVert^2}{v_j}
-d+d\log\frac{v_j}{\sigma^2}
\right).
\]
The available Gaussian KL theorem uses Euclidean coordinates, whereas the SAM
development represents the distributions as affine pushforwards of a standard
Gaussian on a finite-dimensional inner-product space $E$. Reusing the theorem
therefore requires a proved connection between these representations.

A direct attempt exposed a gap in the model: the original bridge allowed an
arbitrary measurable structure on $E$, which did not establish that the
orthonormal coordinate map was a measurable equivalence. Planner diagnosed this
representation mismatch and requested reconstruction. Guided by the source's
Euclidean parameter space, the reconstruction made the compatible measurable
structure explicit in the bridge and its consumers. ProofLoom then constructed
the missing theory: it established the measurable equivalence induced by an
orthonormal basis, identified the affine Gaussian laws with their coordinate
representations by matching means and covariance forms, and proved the transport
of KL divergence across this equivalence.

These constructions made the existing Gaussian KL formula applicable. Combining
it with preservation of the squared norm yielded the target identity. Judge
checked the revised representation and transport bridges against the
source-facing KL statements, and Lean checked the resulting proofs. The
selected-Gaussian prior-grid argument then consumed the proved KL result at its
chosen positive scale. The example shows how a target proof obligation drives
interface reconstruction and the construction of supporting theory before proof
execution can continue. Appendix~\ref{app:sam-construction} provides the Lean
declarations and further implementation details.

\section{SOptLib: Learning from Algorithm Proofs}
\label{sec:assets}

The Learn stage develops SOptLib from completed algorithm proofs. The library contains
verified Lean definitions and lemmas, together with natural-language records of their
construction. These records describe modeling decisions, failed proof routes, revised
approaches, and conditions for reuse. They retain source references and audit evidence,
including evidence for corrected statements and checked counterexamples. Completed
algorithm developments provide worked examples of the resulting theory.
Table~\ref{tab:layers} summarizes the library's mathematical organization.

\begin{table}[htbp]
\caption{Mathematical organization of SOptLib with representative examples.}
\label{tab:layers}
\centering
\setlength{\tabcolsep}{4pt}
\renewcommand{\arraystretch}{1.12}\hyphenpenalty=10000\exhyphenpenalty=10000
\begin{tabular}{@{}>{\RaggedRight\arraybackslash}p{1.3cm} >{\RaggedRight\arraybackslash}p{3.1cm} >{\RaggedRight\arraybackslash}p{\dimexpr\linewidth-4.4cm-4\tabcolsep\relax}@{}}
\toprule
\tablehead{Layer} & \tablehead{Role} & \tablehead{Representative example} \\
\midrule
\textsc{Glue} & Mathlib bridges &
\(\mathbb{E}[X\mid\mathcal{F}]=Y\ \text{a.s.}\ \Rightarrow\ \mathbb{E}[X]=\mathbb{E}[Y]\)\newline
Integrating a conditional identity \\[3pt]
\textsc{Model} & Shared objects &
\(V_h(x,y)=h(y)-h(x)-\langle\nabla h(x),y-x\rangle\)\newline
Bregman divergence \\[3pt]
\textsc{Layer\,0} & Problem properties &
\(\mathbb{E}\lVert\bar\delta_m\rVert^2\le\sigma^2/m\)\newline
Second-moment bound for averaged oracle noise \\[3pt]
\textsc{Layer\,1} & Convergence lemmas &
\(\gamma\langle g,x^+-u\rangle\le V_h(x,u)-V_h(x,x^+)-V_h(x^+,u)\)\newline
Proximal three-point inequality \\
\bottomrule
\end{tabular}
\par\vspace{3pt}
\begin{minipage}{\linewidth}

For the batch bound, \(\bar\delta_m=m^{-1}\sum_{i=1}^m\delta_i\), \(m>0\), and
\(\mathbb{E}\lVert\sum_{i=1}^m\delta_i\rVert^2\le m\sigma^2\).
The point \(x^+\) is a Bregman proximal update, and \(u\) is a feasible comparator.
\end{minipage}
\end{table}

After certification (Section~\ref{sec:topdown:stages}), extraction examines proof
dependencies and construction records to identify reusable mathematical objects and
arguments. Existing Mathlib and SOptLib declarations are reused where
applicable. Generalization replaces algorithm-specific objects with parameters, states
the required assumptions, and reorganizes supporting declarations into independent
interfaces. Extracted definitions are accompanied by lemmas characterizing their
properties. The generalized results are proved in Lean and instantiated in the original
development, with their hypotheses discharged from its existing assumptions. The original
theorem statements and definition types are preserved, and the complete development is
rebuilt to check its use of the library. Independent review checks the mathematical
purpose, generality, and overlap of the declarations, and the evidence and
applicability of the construction records. Feedback guides revisions before merging.

For subsequent algorithms, agents use natural-language directories and symbol search to
locate relevant declarations and construction records. They inspect signatures,
hypotheses, and dependencies before instantiating a result. The associated records guide
modeling and proof planning; for example, they explain which sampling relation supports
a conditional identity and which oracle representation makes it available. New proof
obligations motivate further additions to SOptLib through the same process.
Section~\ref{sec:results} reports library growth, reuse, and source findings.


\section{Experiments and Findings}
\label{sec:results}

\subsection{Experimental Setup}

\noindent\textbf{Tasks and systems.}
We evaluate ProofLoom on fifteen optimization tasks and examine the
mathematical infrastructure produced across \NAlgos{} algorithm developments.
The benchmark contains ten algorithms from Lan's monograph
\citep{lan2020foml} (FOML) and five research-paper algorithms: AMSGrad, STORM,
SPIDER, PAGE, and Sharpness-Aware Minimization (SAM). The research-paper tasks
cover online regret bounds, convergence guarantees for stochastic and finite-sum
nonconvex optimization, and PAC-Bayes generalization bounds.
Each task provides the published algorithm, its assumptions,
a target convergence, regret, or generalization result, and the supporting source
proof. The task is to construct the algorithm definitions, theorem statement,
and proof in Lean. We compare ProofLoom with Trellis, LeanMarathon, Archon,
OpenGauss, Raw Codex, and Raw Codex (Goals). The evaluated ProofLoom system includes
SOptLib, which accumulates reusable definitions, lemmas, and construction
records during the FOML developments. All five research-paper tasks use the same SOptLib snapshot containing only
these FOML results and records, held fixed throughout evaluation.
Each system--task pair has a cumulative generation budget of 48 hours,
including resumed execution. Appendix~\ref{app:baseline-execution} details the
implementations and execution settings.

\noindent\textbf{Human and model-based evaluation.}
Human evaluation follows a 1--7 rubric jointly established by two Lean experts
(Table~\ref{tab:human-rubric}), with independent blind human and LLM ratings
and disagreements adjudicated by a second human expert. We also use GPT, Gemini,
DeepSeek, and Claude to evaluate artifacts with G-Eval \citep{geval2023},
adapted to complete algorithm formalizations, and FidelityEval, a source-guided
evaluation skill for Lean formalizations.
These two protocols each produce 0--100 scores. Table~\ref{tab:main-results}
reports task-averaged scores for each source group, protocol, and judge.
Evaluation procedures are provided in Appendix~\ref{app:evaluation}.

\subsection{Main Results}

\begin{table}[!t]
\caption{\textbf{Seven-system comparison across human and model-based evaluation.}
Human ratings use a 1--7 scale; G-Eval and FidelityEval scores use a 0--100 scale (higher is better).}
\label{tab:family-joint}\label{tab:main-results}
\centering
\begingroup
\setlength{\tabcolsep}{4pt}\renewcommand{\arraystretch}{1.0}
\begin{tabular}{@{}c l c @ {\hspace{3pt}} rrrr @ {\hspace{3pt}} rrrr@{}}
\toprule
& & \tablehead{Human} & \multicolumn{4}{c}{\tablehead{G-Eval}} & \multicolumn{4}{c}{\tablehead{FidelityEval}} \\
\cmidrule(lr){3-3}\cmidrule(lr){4-7}\cmidrule(l){8-11}
& \tablehead{System} & (1--7) $\uparrow$ & \shortstack{GPT\\5.6-\\sol} & \shortstack{Gemini\\3.8\\Flash} & \shortstack{DeepSeek\\V4\\Pro} & \shortstack{Claude\\Opus\\5} & \shortstack{GPT\\5.6-\\sol} & \shortstack{Gemini\\3.8\\Flash} & \shortstack{DeepSeek\\V4\\Pro} & \shortstack{Claude\\Opus\\5} \\
\midrule
& Raw Codex & 2.8 & 36.5 & 36.7 & 37.6 & 39.9 & 31.4 & 28.4 & 34.3 & 29.5 \\
& Raw Codex (Goals) & 3.0 & 35.3 & 36.2 & 36.2 & 41.3 & 30.9 & 29.4 & 35.7 & 29.0 \\
\addlinespace[2pt]
& OpenGauss & 3.5 & 43.8 & 48.1 & 47.6 & 48.1 & 43.1 & 35.9 & 44.9 & 36.7 \\
& LeanMarathon & 4.4 & 72.9 & 74.3 & 75.9 & 72.4 & 72.1 & 64.6 & 77.0 & 69.3 \\
& Trellis & 4.5 & 63.1 & 67.2 & 66.5 & 65.6 & 67.0 & 64.9 & 64.8 & 60.9 \\
& Archon & 4.9 & 61.4 & 66.0 & 65.2 & 67.7 & 72.8 & 64.6 & 71.9 & 66.8 \\
\cmidrule(l){2-11}
\rowcolor{tablehighlight}
\cellcolor{white}\multirow[c]{-7}{*}[3pt]{\rotatebox{90}{\bfseries A. FOML}} & \textbf{ProofLoom} & \textbf{6.3} & \textbf{92.0} & \textbf{87.9} & \textbf{88.6} & \textbf{89.4} & \textbf{87.8} & \textbf{93.3} & \textbf{90.1} & \textbf{85.8} \\
\addlinespace[5pt]\midrule
& Raw Codex & 2.8 & 36.3 & 39.0 & 34.6 & 40.8 & 35.3 & 37.2 & 35.0 & 26.2 \\
& Raw Codex (Goals) & 3.0 & 47.9 & 47.6 & 44.8 & 48.0 & 44.7 & 46.2 & 45.4 & 36.6 \\
\addlinespace[2pt]
& OpenGauss & 3.8 & 54.7 & 52.2 & 48.8 & 51.8 & 52.5 & 50.2 & 55.8 & 43.6 \\
& LeanMarathon & 4.2 & 65.5 & 66.8 & 68.8 & 63.8 & 66.9 & 63.4 & 68.8 & 60.6 \\
& Trellis & 4.8 & 64.7 & 76.2 & 66.0 & 67.8 & 67.5 & 71.6 & 69.6 & 65.2 \\
& Archon & 5.0 & 72.7 & 69.2 & 67.6 & 70.8 & 77.1 & 75.0 & 79.2 & 72.4 \\
\cmidrule(l){2-11}
\rowcolor{tablehighlight}
\cellcolor{white}\multirow[c]{-7}{*}[3pt]{\rotatebox{90}{\bfseries B. Papers}} & \textbf{ProofLoom} & \textbf{6.4} & \textbf{89.9} & \textbf{84.6} & \textbf{80.8} & \textbf{90.6} & \textbf{81.5} & \textbf{89.6} & \textbf{83.8} & \textbf{86.6} \\
\bottomrule
\end{tabular}
\endgroup
\vspace{4pt}
\begin{minipage}{\textwidth}
Values are means within each source group; bold indicates column maxima.
\end{minipage}

\end{table}

\noindent\textbf{Human and model-based assessment.}
ProofLoom has the highest human mean in both source groups: 6.3/7 on FOML and
6.4/7 on research-paper tasks, compared with 4.9/7 and 5.0/7 for Archon, the
strongest baseline. The 1.4-point margin in each group reflects higher assessed
source faithfulness and proof completeness across both textbook and
research-paper developments.

ProofLoom also leads in all 16 combinations of source group, evaluation
protocol, and model judge. Relative to the strongest baseline in each column,
its margins range from 12.7 to 28.4 points on FOML and from 4.4 to 19.8 points
on research-paper tasks. With SOptLib fixed after the FOML developments,
the complete ProofLoom system maintains high source-faithfulness scores
across independent research papers while reusing its own definitions,
lemmas, and construction records.

\subsection{Mechanism Ablation}
\label{sec:mechanism-ablation}

We compare the full system with two ablations, removing Judge or Planner--Audit,
on the same 43 cases (129 runs). Table~\ref{tab:mechanism-ablation} reports
crossing and incorrect-repair percentages for modeling, interface, or representation
mismatches (A, 29 cases), and false or unprovable propositions or missing
mathematical assumptions (B, 14 cases). Human reviewers assign case labels using
the main evaluation's
independent assessment and expert-adjudication procedure;
Appendix~\ref{app:ablation-scoring} defines the labels and rates. The labels assess the diagnosis of the root
obstruction and the proposed repair. Crossing records a correct diagnosis and a source-faithful repair
direction. Partial records an unresolved diagnosis or repair direction.
Incorrect records an erroneous replacement or an unsupported change to the
source-facing mathematical statement, supported by explicit evidence.

\noindent\textbf{Judge.}
Removing Judge lowers crossing in group A from 72.4\% to 58.6\% and raises
incorrect repairs from 3.4\% to 17.2\%. In group B, crossing remains at 85.7\%,
while incorrect repairs rise from 0.0\% to 7.1\%. Across the two groups, the
incorrect count rises from one to six. This pattern supports Judge's role in
checking the mathematical validity of candidate repairs.

\noindent\textbf{Planner--Audit.}
Removing Planner--Audit lowers crossing from 72.4\% to 62.1\% in group A and
from 85.7\% to 78.6\% in group B. The total partial count rises from nine to
thirteen, while the incorrect count remains at one. Removing Planner--Audit
leaves more cases partial, indicating less progress through blockers.
The full system crosses 33 cases, compared with
29 for either ablation.

\subsection{Formalization Products and Source Findings}

\begin{figure}[!htbp]
\centering
\newsavebox{\ablationbody}
\newsavebox{\ablationpanel}
\newsavebox{\productcaption}
\newsavebox{\productgraphic}
\newlength{\ablationpanelheight}
\newlength{\productgraphicheight}
\begin{lrbox}{\ablationbody}
\begin{minipage}[t]{0.50\textwidth}
\centering

\begingroup

\setlength{\tabcolsep}{1.5pt}
\renewcommand{\arraystretch}{1.18}
\begin{tabular*}{\linewidth}{@{\extracolsep{\fill}}lcc@{}}
\toprule
\tablehead{Configuration} & Crossing\,$\uparrow$ & Incorrect\,$\downarrow$ \\
\midrule
\multicolumn{3}{@{}p{\linewidth}@{}}{\textbf{A. Model/interface mismatch} ($n=29$)} \\
\addlinespace[2pt]
w/o Judge & 58.6 & 17.2 \\
w/o Planner--Audit & 62.1 & \textbf{3.4} \\
\midrule
\rowcolor{tablehighlight}
\textbf{ProofLoom} & \textbf{72.4} & \textbf{3.4} \\
\addlinespace[4pt]
\midrule
\multicolumn{3}{@{}p{\linewidth}@{}}{\textbf{B. False/underspecified claims} ($n=14$)} \\
\addlinespace[2pt]
w/o Judge & \textbf{85.7} & 7.1 \\
w/o Planner--Audit & 78.6 & \textbf{0.0} \\
\midrule
\rowcolor{tablehighlight}
\textbf{ProofLoom} & \textbf{85.7} & \textbf{0.0} \\
\bottomrule
\end{tabular*}
\endgroup

\end{minipage}
\end{lrbox}
\begin{lrbox}{\ablationpanel}
\begin{minipage}[t]{0.50\textwidth}
\captionof{table}{Corner-case ablation. Crossing and incorrect repairs (\%).}
\label{tab:mechanism-ablation}
\centering
\usebox{\ablationbody}\par
\raggedright
Remaining cases are partial.
\end{minipage}
\end{lrbox}
\begin{lrbox}{\productcaption}
\begin{minipage}[t]{0.45\textwidth}
\captionof{figure}{Code and reuse in 33 construction snapshots.}
\label{fig:formalization-scale}
\end{minipage}
\end{lrbox}
\setlength{\ablationpanelheight}{\dimexpr\ht\ablationpanel+\dp\ablationpanel\relax}
\setlength{\productgraphicheight}{\dimexpr\ablationpanelheight-\abovecaptionskip-\ht\productcaption-\dp\productcaption\relax}
\sbox{\productgraphic}{%
\includegraphics[height=\productgraphicheight,width=0.45\textwidth,keepaspectratio,trim=23.166bp 18.207bp 30.399bp 17.763bp,clip]{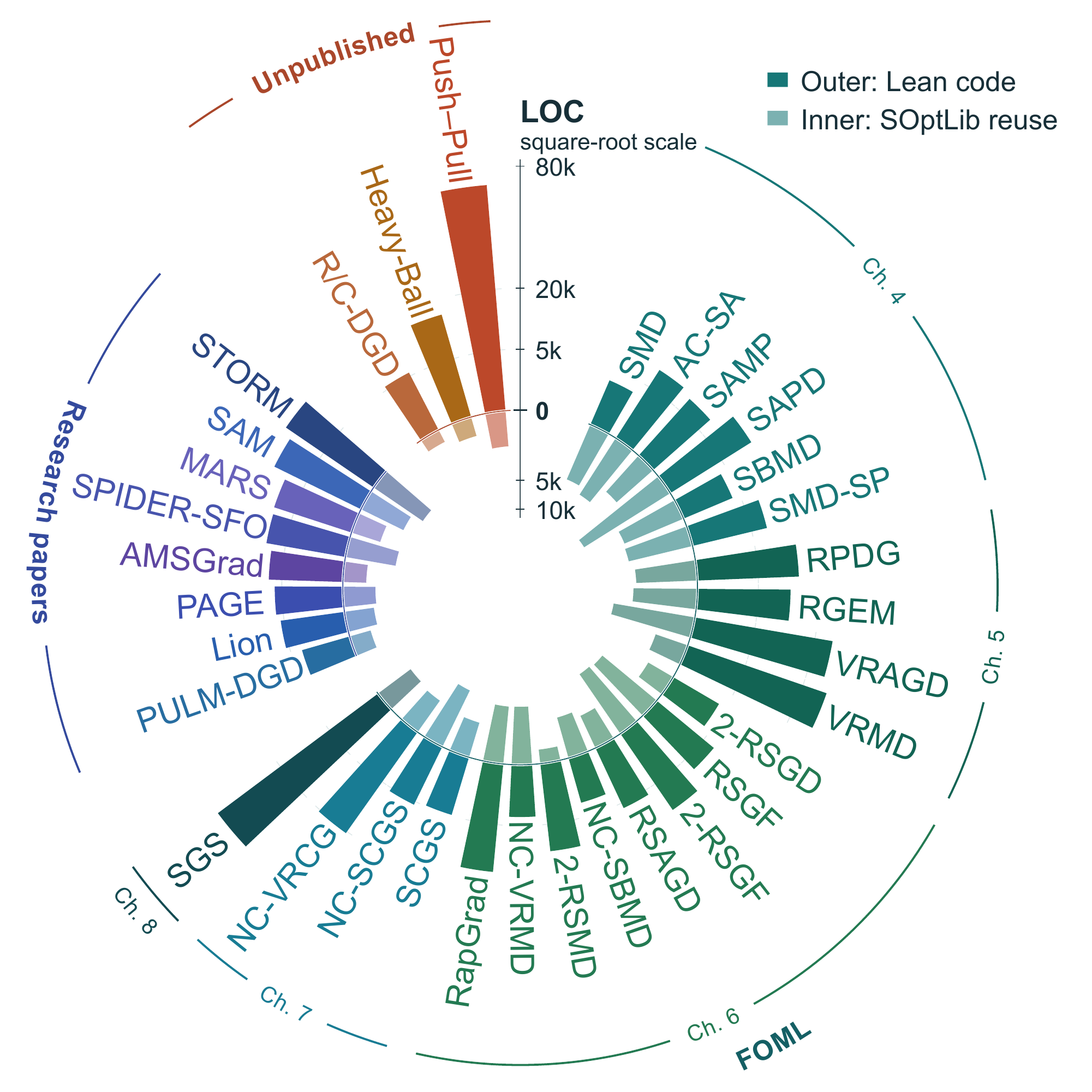}}
\noindent
\vtop{\vskip0pt\hbox{\usebox{\ablationpanel}}}\hfill
\vtop{\vskip0pt
\vbox to\ablationpanelheight{%
\offinterlineskip
\vskip\abovecaptionskip
\hbox to0.45\textwidth{\hfil\usebox{\productgraphic}\hfil}%
\vfil
\hbox{\usebox{\productcaption}}\kern0pt}}
\end{figure}

The census covers \NAlgos{} developments: 22 from FOML, eight from research
papers including MARS~\citep{yuan2025mars}, and three from research notes.
It includes all fifteen benchmark tasks and one unreleased development
(Appendix~\ref{app:corpus-census}). Figure~\ref{fig:formalization-scale} shows
algorithm code (outer bars) and reused SOptLib support (inner bars).
Appendix~\ref{app:sam-construction} details SAM's Gaussian KL construction.

\noindent\textbf{Source findings.} Across the full audit, formalization identifies
\NSourceIssues{} independent discrepancies affecting
\NSourceIssueAlgorithms{} developments. They include incorrect formulas, proof gaps,
and algorithm--analysis mismatches. Appendix~\ref{app:source-findings} records their locations, evidence, and repair scopes.

\paragraph{Finding 1: a Gaussian-scale mismatch in SAM's generalization proof.}
SAM's Theorem~2 assumes that Gaussian perturbations at scale $\rho$ do not
decrease the population loss \citep{foret2021sam}. To concentrate perturbations
inside the radius-$\rho$ neighborhood, its proof selects the smaller standard
deviation $\sigma=\rho/[\sqrt{k}(1+\sqrt{\log(n)/k})]$, where $k$ is the parameter
dimension and $n$ the sample size, then invokes the same premise at $\sigma$.
This change of scale need not preserve the inequality. For the smooth bounded
loss $\ell(x)=0.5-0.2\cos x+0.1\cos(2x)$, we have $\ell(0)=0.4$; with $k=1$,
$n=16$, and $\rho=2$, the Gaussian averages are approximately $0.473$ at
$\rho$ and $0.382$ at $\sigma$. Thus the stated premise holds while the premise
needed by the proof fails. The corrected formulation explicitly requires
nondecrease at the selected scale and uses that scale consistently throughout
the Gaussian argument (Appendix~\ref{case:source-A17}).

\paragraph{Finding 2: a textbook proof that steps outside its domain.}
Lan assumes a closed convex feasible set $X$ and smoothness only on $X$, yet
the proof of Lemma~5.8 evaluates a potential at $x-\nabla\varphi(x)/L$, which
may leave $X$ \citep{lan2020foml}. The required bound is
$\|g_i(x)-g_i(y)\|^2/(2L_i)\le
f_i(x)-f_i(y)-\langle g_i(y),x-y\rangle$, where $g_i$ denotes the intrinsic
component gradient. The agent automatically formalizes open-domain
Baillon--Haddad cocoercivity on $U=\operatorname{ri}(X)$, open in the affine
span of $X$, following \citet[Theorem~3.1]{perezarosvilches}.
Under the additional assumption that each $g_i(U)$ is convex, it derives
the one-sided bound using \citet[Lemma~3.3]{wachsmuth2022open}, extends it to
$X$ by relative-interior density and continuity, and completes the Lean
proofs connecting Lemma~5.12 to the VRMD convergence results.

\section{Related Work}
\label{sec:related}

\textbf{Formalizing optimization and its foundations.} Mathlib, Coquelicot, and
Isabelle provide analysis and probability foundations
\citep{mathlib,boldo2015coquelicot,avigad2017clt}. OptLib formalizes convex analysis
and first-order optimization rates \citep{wen2024firstorder}.
Other developments cover bounded-noise SGD
\citep{cassie2026sgd}, and almost-sure reinforcement-learning convergence under
Markovian sampling \citep{zhang2025rl}.

\textbf{Research-level autoformalization and faithfulness.} Autoformalization covers statements and projects
\citep{autoformalization_survey2025}. MASA uses multi-agent generation
\citep{masa2025}; Aria retrieves and synthesizes definitions through dependency graphs
\citep{wang2026aria}; ReForm iteratively revises statement semantics
\citep{reform2025}. M2F demonstrates large-scale textbook formalization with explicit dependencies
among source declarations \citep{wang2026m2f}. Archon formalizes
research-level proofs through autonomous gap filling and refactoring
\citep{ju2026automated}. Open Gauss supports project-level workflows
\citep{mathinc2026opengauss}; LeanMarathon and Trellis organize long-horizon
proof development \citep{zhang2026leanmarathon,trellis2026}. ProofFlow formalizes source proofs
through dependent lemmas \citep{cabral2026proofflow}.
Semantic assessment includes learned critics and alignment models
\citep{criticlean2025,lu2025formalalign}, statement-equivalence checks
\citep{faithfulmetric2025,li2024symbolic}, and roundtrip verification and repair for legal rules
\citep{amrollahi2026roundtrip}.
\citet{dollmsgame2026} document axiom fabrication and premise mistranslation. ProofLoom constructs theory from convergence-proof obligations; signature
contracts and Planner--Audit govern source-faithful model and interface revisions.

\textbf{Neural theorem proving and proof search.} LeanDojo supports
retrieval-augmented proving \citep{yang2023leandojo}; LeanSearch v2 studies global
premise retrieval for complete proofs \citep{gao2026leansearchv2}. DeepSeek-Prover trains on
synthetic formal proofs \citep{deepseekprover2024}; AlphaProof combines learning
and proof search \citep{alphaproof2024}; Lean Copilot assists interactive proof
development \citep{song2024leancopilot}. miniF2F evaluates
completion of given formal goals \citep{zheng2022minif2f,ospanov2025minif2f}. LEGO-Prover builds
reusable lemma libraries during proof search \citep{wang2023legoprover},
and LeanAgent studies continual learning across Lean repositories
\citep{kumarappan2025leanagent}. SOptLib retains reusable theory and construction records.

\section{Limitations}
\label{sec:limitations}
Semantic review relies on LLM judgments; repeated errors
can shift assumptions or conclusions away from the source. Construction, semantic
checks, and audit records require additional model calls and Lean checks;
research-level formalization can take hours. Our evaluation covers textbook and
research-paper tasks in stochastic optimization. Prompts and control flow were
initially developed on a small FOML subset, with infrastructure bugs fixed in
subsequent iterations. \textsf{reconstruct} and \textsf{prove} use source proofs to identify
intermediate obligations. Applicability to more abstract domains and less detailed
proofs requires further evaluation.

\section{Conclusion}
\label{sec:conclusion}
ProofLoom develops machine-checked formalizations of stochastic optimization
algorithms through proof-obligation-driven theory construction. Signature contracts
and Planner--Audit make reconstruction decisions and proof dependencies available
for source-grounded review, while SOptLib accumulates verified mathematical
infrastructure and construction knowledge for subsequent developments. Across
textbook and research-paper tasks, human and multi-model evaluations show higher
average source-faithfulness scores than the evaluated baselines. The developments
also expose \NSourceIssues{} incorrect formulas, proof gaps, and algorithm--analysis mismatches
across \NSourceIssueAlgorithms{} developments. These results show how algorithm formalization can build reusable foundations
and provide checkable evidence for examining and refining the literature.


\clearpage
\appendix

\paragraph{Appendix guide.}
Appendix~\ref{app:evaluation} specifies the experimental setup and evaluation
protocols and corpus census. Appendix~\ref{app:source-findings} presents the mathematical
findings and their evidence. Appendix~\ref{app:implementation} gives the system
implementation and a worked example; Appendix~\ref{app:soptlib-experiments}
describes SOptLib construction, reuse, and ablation.

\section{Evaluation Protocols and Experimental Setup}
\label{app:evaluation}

\subsection{Task Scope and Resources}

\paragraph{Reporting boundary.} The comparison contains \NBenchAlgos{} source
tasks: ten FOML algorithms (SMD, SBMD, SCGS, SAPD, NSAGD, SNCCG, RPDG,
SCCSP, NSMD, and SZO) and AMSGrad, STORM, SPIDER, PAGE, and Sharpness-Aware
Minimization (SAM). Each task is evaluated using seven anonymized candidate
artifacts. Assessments are included only when they match the designated task
and artifact versions and preserve candidate anonymity.
The formalization task starts from the source package and a fresh
algorithm-specific target project; assessment covers the algorithm and its main
result. ProofLoom's FOML tasks access the SOptLib support available at their
starting stage; its research-paper tasks share a fixed support version containing only
FOML-derived definitions and lemmas, held fixed throughout evaluation
(Appendix~\ref{app:soptlib-experiments}).

\subsection{Baseline Implementations and Execution Settings}
\label{app:baseline-execution}

\paragraph{Implementations and inputs.}
We use the official implementations of Trellis, LeanMarathon, Archon, and
OpenGauss at the revisions in Table~\ref{tab:baseline-implementations}.
Each task supplies the published source, target specification, and Lean
environment. Adapters organize these materials for each system's native
workflow. ProofLoom's FOML tasks use the indexed SOptLib definitions and lemmas
available before each task. Its five research-paper
tasks use the same fixed SOptLib version, containing only definitions and lemmas
from FOML developments.

\begin{table}[!htbp]
\caption{Baseline repository revisions and benchmark adaptations.
OpenGauss uses \texttt{lean4-skills} revision \texttt{e63d0723bbc9}.}
\label{tab:baseline-implementations}
\centering
\setlength{\tabcolsep}{4pt}
\begin{tabular}{@{}>{\RaggedRight\arraybackslash}p{0.17\linewidth}>{\RaggedRight\arraybackslash}p{0.20\linewidth}>{\RaggedRight\arraybackslash}p{0.57\linewidth}@{}}
\toprule
System & Revision & Workflow and runtime adaptation \\
\midrule
Trellis & \texttt{723bf99d0834} & Native setup and proof workflow; source TeX and target labels are prepared as reference inputs. \\
LeanMarathon & \texttt{9ace81e67b55} & Blueprinter, Target-Reviewer/Refiner, and Worker/Refiner stages; local Git and issue tracking support execution. \\
Archon & \texttt{5e9ae7615efa} & Initialization, dependency graph construction, and the plan--prove--review loop, starting from a blueprint scaffold. \\
OpenGauss & \texttt{f87633900ae1} & \texttt{/autoformalize} with source-coverage and verification instructions; noninteractive Codex execution replaces the terminal interface. \\
\bottomrule
\end{tabular}

\end{table}

\paragraph{Configuration and execution budget.}
ProofLoom and all baselines use GPT-5.5 (\texttt{gpt-5.5}) through Codex
with \texttt{high} reasoning effort for generation. This setting applies
throughout the ProofLoom pipeline and to each baseline, including Archon.
We retain each system's default workflow settings, with the model,
budget, input and runtime adaptations described here. Each system--task pair
has a cumulative generation limit of 48 hours, including resumed execution.
Human intervention is limited to repairing infrastructure bugs. Approval
gates in human-in-the-loop workflows are automatically approved.
Runtime maintenance covers dependencies, caches, and tool availability.

\paragraph{Direct generation baselines.}
Raw Codex and Raw Codex (Goals) receive the same task prompt in separate workspaces.
Raw Codex uses direct Codex execution; Raw Codex (Goals) uses the built-in Goals
workflow. Both use the generation model, reasoning effort, and time limit
specified above. Their workspaces retain the generated Lean sources,
dependency manifests, and build-status files.

\subsection{Human Evaluation and Adjudication}

\paragraph{Human evaluation and expert adjudication.}
Two Lean experts jointly established the seven-point rubric through discussion,
specifying the criteria for each level and the score caps for substantive defects
(Table~\ref{tab:human-rubric}).
For each task $a$ and system $m$, the first human reviewer $H_1$ and an
LLM reviewer $J$ independently assess the same anonymized Lean artifact,
source, and verification evidence. Reviewer $J$ uses \texttt{gpt-5.6-sol}
through Codex with \texttt{xhigh} reasoning effort. Both reviewers return integer
ratings $h_{a,m}$ and $\ell_{a,m}$ on the 1--7 scale, together with their reasons.
Agreement preserves the first human rating. For a disagreement, a second
human expert $H_2$ reviews the evidence and both assessments, reports a
rationale, and assigns an adjudicated integer rating $z_{a,m}$ on the same
scale. The final rating is
\[
q_{a,m}=\begin{cases}
h_{a,m}, & h_{a,m}=\ell_{a,m},\\
z_{a,m}, & h_{a,m}\ne\ell_{a,m}.
\end{cases}
\]
The independent LLM assessment provides an additional check for possible
oversights, while concentrating the second expert's review on disagreements
makes efficient use of limited expert time.
This procedure yields \NHumanRatings{} final ratings for fifteen tasks and
seven systems. The auxiliary LLM assessment uses the same seven-level scale;
G-Eval and FidelityEval are reported separately on their 0--100 scales.

\begingroup
\renewcommand{\arraystretch}{1.12}
\setlength{\LTcapwidth}{\textwidth}
\begin{longtable}{@{}cp{0.88\linewidth}@{}}
\caption{Shared rubric for the independent human and LLM ratings and expert
adjudication. Reviewers assign the highest level whose conditions are satisfied,
then apply the score caps described below.}
\label{tab:human-rubric}\\
\toprule
\tablehead{Score} & \tablehead{Criteria}\\
\midrule\endfirsthead
\multicolumn{2}{@{}l}{\tablename\ \thetable\ (continued)}\\
\toprule
\tablehead{Score} & \tablehead{Criteria}\\
\midrule\endhead
\midrule\multicolumn{2}{r@{}}{Continued on next page}\\\endfoot
\bottomrule\endlastfoot
1 & Missing, unrelated, or plainly wrong formalization of the target algorithm or source result.\\
2 & Names, definitions, comments, or a theorem shell are present; the algorithm update or target conclusion is absent, or the endpoint relies on an unverified placeholder.\\
3 & Some algorithm objects and algebra are present, while a core semantic component is missing and the central derivation remains incomplete.\\
4 & The algorithm and endpoint follow the source structure and compile, while a central bridge is assumed or opaque, or the update, randomness, quantifiers, or conclusion is materially simplified or changed.\\
5 & The algorithm, source assumptions, and endpoint are substantially recognizable, and most of the proof chain is checked; a central closure or alignment condition remains unresolved or explicitly retained as a proof boundary.\\
6 & The algorithm and endpoint are faithful, and the core chain is checked: update, one-step inequality, telescoping/output, stochastic cancellation or variance, and final bound. The endpoint uses accepted standard foundations. Remaining issues are local scope restrictions, visible specializations, domain regularity caveats, or a small number of noncentral hidden source assumptions.\\
7 & All level-6 conditions hold over the complete source scope. Every source premise is mapped, derived obligations are proved, the endpoint matches the stated generality or an explicitly disclosed equivalent correction, and build and source-alignment checks are clean.\\
\end{longtable}
\endgroup

\paragraph{Rubric application.}
A missing or unrelated source endpoint caps the rating at 1; materially wrong
algorithm semantics or a wrong target conclusion caps it at 3; an assumed
central proof bridge caps it at 5; and partial source scope or remaining
hidden source assumptions caps it at 6. An explicitly identified source error
and its mathematically justified correction receive credit according to the
same rubric. For a corrected source result, level 7 requires the original claim,
its defect, the corrected
claim, their relationship, and a complete proof of the corrected conclusion.

\begin{table}[!htbp]
\caption{Evaluation protocols. Human ratings and the two model-based protocols retain separate scales and aggregates.}
\label{tab:evaluation-protocol}
\centering
\setlength{\tabcolsep}{4pt}
\renewcommand{\arraystretch}{1.15}
\begin{tabular}{@{}lccc@{}}
\toprule
\tablehead{Protocol} & \tablehead{Input} & \tablehead{Primary endpoint} & \tablehead{Role}\\
\midrule
Human review & Artifact + source & 1--7 adjudicated rating & Artifact quality \\
G-Eval & Seven candidates + source & 0--100 joint score & Artifact quality \\
FidelityEval & Seven candidates + source & 0--100 joint score & Artifact quality \\
\bottomrule
\end{tabular}

\end{table}

\subsection{Model-Based Evaluation and Aggregation}

\paragraph{Model judges and harnesses.} Both automatic protocols are evaluated
with \texttt{gpt-5.6-sol} (GPT in the tables), \texttt{gemini-3.8-flash},
\texttt{deepseek-v4-pro}, and \texttt{claude-opus-5}. The respective execution
interfaces are Codex, the Gemini harness, the DeepSeek harness, and Claude Code;
GPT uses \texttt{xhigh} reasoning effort; Claude uses \texttt{high} effort with a
one-million-token context window. Model identity is distinct from the
execution harness. Each joint assessment receives the source package and seven
anonymized candidates; stored mappings recover system identities after scoring.
Inputs, instructions, model configuration, raw verdicts, and execution receipts
are archived. Since execution interfaces also differ, cross-judge comparisons
characterize these evaluator configurations as a whole.

\paragraph{Scoring setup.} Our G-Eval adaptation uses LLM-generated evaluation
steps fixed before scoring to guide source-to-Lean review. Judges return
structured component scores with supporting evidence. Our task-specific rubric
allocates 25 points each to algorithm fidelity, source alignment, theorem
derivation, and verification. Judges are instructed to sum the component
scores into a 0--100 total. FidelityEval combines
its source-guided inspection workflow with the joint-evaluation rubric used
for this benchmark: algorithm contract (40), main theorem or refutation (40),
verification and trust (15), and public coverage and integration (5). Both
protocols return a separate 0--100 score for each candidate.

\paragraph{Verification adjustment and totals.}
Eligibility for verification credit requires successful compilation of the
designated files at their specified hashes, zero detected proof placeholders
and associated compiler warnings, and completed proof-integrity checks
for the evaluated artifact. Eligible artifacts receive full verification credit:
25 points for G-Eval and 15 for FidelityEval. Other artifacts retain the
judge's verification score. The final total changes by exactly this component's
adjustment. Aggregation uses the reported total plus the verification adjustment.
Source-alignment and mathematical judgments retain their original scores.

\paragraph{FidelityEval construction and workflow.}
FidelityEval is generated from a human-authored meta-prompt and a collection
of studies on mathematical formalization, autoformalization, and evaluation.
The literature includes \emph{Beyond Compilation} \citep{zhang2026beyondcompilation}, \emph{CriticLean}
\citep{criticlean2025},
\emph{The Faithfulness Gap} (BPF) \citep{mohammad2026faithfulnessgap}, and
\emph{FormalRx} \citep{wang2026formalrx}, together with studies
of model-based evaluation and rubric design \citep{geval2023}. The meta-prompt specifies the
evaluation objective and source materials, leaving the choice of formalization
techniques and inspection procedures to the skill generator. Generator and
auditor instructions treat candidate systems symmetrically; system identities,
benchmark outputs, previous scores, and desired rankings are withheld during
construction.

The skill generator produces a candidate evaluation procedure. A separate
adversarial auditor examines it using the same task objective and literature,
develops counterexamples and boundary cases, and returns concrete revision
requests. The generator incorporates this feedback through ten rounds of
automated adversarial refinement to produce the evaluation skill.
During scoring, FidelityEval identifies the source algorithm, assumptions,
target conclusion, and proof obligations; it then inspects the corresponding
Lean definitions, statements, and proof dependencies. The resulting assessment
uses the joint-evaluation rubric above and returns supporting source and Lean
evidence.

\paragraph{Assessment counts and aggregation.}
For each task and each 0--100 protocol, Codex produces three assessments;
Gemini, DeepSeek, and Claude produce one each. This allocation divides the
evaluation budget between repeated assessment with GPT and comparison across
four evaluator configurations. GPT repetitions provide observations of
within-evaluator variation; the other configurations broaden the comparison
across judges. Each joint assessment scores
seven anonymized candidates. Across fifteen tasks and two protocols, this
gives $15\times2\times(3+1+1+1)=\NJointAssessments{}$ joint assessments and
$\NJointAssessments{}\times7=\NModelRatings{}$ candidate ratings.

For task $a$, system $m$, protocol $e$, and judge $j$, let $s_{a,m,e,j,r}$
denote assessment $r$ and $n_{a,e,j}$ its repetition count. For source group
$\mathcal A_g$, the task mean and group mean are
\[
\bar s_{a,m,e,j}=\frac{1}{n_{a,e,j}}
\sum_{r=1}^{n_{a,e,j}}s_{a,m,e,j,r},
\qquad
S_{g,m,e,j}=\frac{1}{|\mathcal A_g|}
\sum_{a\in\mathcal A_g}\bar s_{a,m,e,j}.
\]
Human group means use the final adjudicated ratings:
\[
H_{g,m}=\frac{1}{|\mathcal A_g|}\sum_{a\in\mathcal A_g}q_{a,m}.
\]
Each group mean covers all tasks in its source group. Scores are reported
separately by judge, protocol, and scale.

\subsection{Mechanism Ablation Protocol}

\paragraph{Case selection from formalization audits.}
We selected 43 cases from formalization audits by screening
obstruction descriptions, proposed changes to definitions, interfaces, or
assumptions, and source-alignment judgments. Each case includes its algorithm,
intervention round, and supporting source and Lean evidence. The cases cover
model and interface mismatches, false claims, and missing assumptions.
Each obstruction is evaluated under the full system and both component
ablations.

\paragraph{Purpose of the case-level comparison.}
Planner--Audit and Judge jointly govern reconstruction through planning,
auditing, and review of proposed revisions. Their decisions shape the
development states and proof obligations encountered later in a run.
We evaluate the Planner--Audit and signature-contract parts of this loop at
selected obstructions rather than by removing them from complete end-to-end
runs. Each case starts from the same development state, so the
comparison isolates diagnosis, repair direction, and source alignment.
End-to-end endpoint quality also depends on whether all downstream obligations
close within the available budget; an unresolved interface or contract corner
case can otherwise dominate the endpoint and obscure the mechanism's
contribution. The full system is evaluated end to end, while the case-level
study is used for mechanism attribution.
Running each configuration from the same development state and source
material makes the immediate mathematical obstruction common across
configurations. We examine how removing either component changes the diagnosis,
the proposed repair, and progress toward resolving that obstruction. This
design concentrates the evaluation budget on the decisions these components
govern, while the continuations reveal how each configuration responds
to the same difficulty. Full-task evaluations measure the combined outcome
of successive decisions; this comparison examines repair behavior at the
selected obstructions.

\paragraph{Execution and component removal.}
Runs enter \textsf{reconstruct} from the development and source materials at
the obstruction. Removing Judge retains Planner and Audit, with
Refactor controlling iterations. Removing Planner--Audit retains Judge, uses
route selection and a free-form plan, and disables runtime audits and dedicated
infrastructure routing. All configurations retain compilation, input-integrity,
and protected-signature checks. Initial runs receive two hours per case and
configuration. Each continuation uses the budget assigned to its case and
configuration.

\paragraph{Ablation labels and rates.}
\label{app:ablation-scoring}
The ablation uses the same independent human and LLM assessment procedure,
with disagreements resolved by a second expert. Reviewers apply the three
semantic label definitions to each case and configuration. Crossing requires
an identified root obstruction and a source-faithful repair direction, with
subsequent proof obligations allowed. Incorrect requires explicit evidence
of a wrong repair or an unsupported change to the source-facing mathematical
statement. The remaining cases receive partial. Reviewers report the source
and Lean evidence and the reason for each label.

Let $y_{i,c}$ be the final label for case $i$ under configuration $c$, and let
$\mathcal I_g$ be the case set in group $g$. The reported percentages are
\[
\begin{aligned}
\operatorname{Crossing}_{g,c}
&=\frac{100}{|\mathcal I_g|}
\sum_{i\in\mathcal I_g}\mathbf{1}\{y_{i,c}=\mathrm{crossing}\},\\
\operatorname{Incorrect}_{g,c}
&=\frac{100}{|\mathcal I_g|}
\sum_{i\in\mathcal I_g}\mathbf{1}\{y_{i,c}=\mathrm{incorrect}\}.
\end{aligned}
\]
The same cases are assessed under all three configurations, with
$|\mathcal I_A|=29$ and $|\mathcal I_B|=14$.

\paragraph{Case study: RAPP gradient-memory initialization.}
The RAPP gradient-memory case illustrates two distinct failure
modes at the same obstruction. The residual estimate in Lemma~6.13 of
\citet{lan2020foml} requires the componentwise identity
$y_i^t=\nabla\psi_i(x_i^t)$ for a fixed subproblem. Algorithm~6.9 refreshes
the sampled component and retains the other components. Propagating this
identity therefore requires the initial relation
$y_i^0=\nabla\psi_i(x_i^0)$ for every $i$; the formal interface used in the case
allowed arbitrary initial gradient memory.

The full system made this initialization condition explicit, proved the
all-component invariant by induction over sampled and retained components,
and established its transfer through the generated outer recursion using the
corresponding sample blocks. This repaired the local memory boundary and
received a crossing label.
Without Judge, the run also constructed memory lemmas, but strengthened the
public outer-domain predicate with a requirement that every preceding
subproblem satisfy the source-domain conditions at its actual sample offset.
The revised interface assumed this additional restriction without deriving
it from the existing source assumptions, leading to an incorrect label.
Without Planner--Audit, the run identified the domain mismatch but retained
arbitrary initial gradient memory and supplied no inductive memory invariant,
leading to a partial label.
The outcomes distinguish an unsupported restriction of the theorem's applicability
from an unfinished invariant construction. They illustrate the complementary
roles of reviewing source-facing changes and organizing the proof obligations
needed for a repair. The crossing label records the repaired memory boundary;
subsequent convergence-proof obligations remain separate.

\subsection{Validation and Reproducibility}

\paragraph{Output validation.}
Checks cover input identity, execution and reading evidence, candidate coverage,
scoring dimensions, and a recoverable final output.
Arithmetic discrepancies and verification adjustments follow the scoring rules above.
Aggregates use
completed assessments of the designated task and artifact versions, with
provenance attached to each assessment.

\paragraph{Integrity and invalid runs.} Build validity, target dependency integrity, and raw
\lean{sorry} counts are reported separately from semantic faithfulness. A target is marked
sorry-free after checking for zero proof placeholders throughout its dependency
closure; the raw number of sorry sites remains a descriptive statistic. Authentication, transport, corrupted-input failures, and unfinished judge outputs
are marked invalid or incomplete. Candidate-generation timeouts after valid work
remain scored as incomplete artifacts. System rankings use artifact-quality scores.

\paragraph{Reproduction package.}
The accompanying package is indexed by
\mbox{\path{docs/PAPER_INDEX.md}}; raw scores and case labels are listed in
\mbox{\path{experiments/README.md}}. Candidate and development manifests identify the
versions used for evaluation and the released Lean sources.
Each project's \mbox{\path{lean-toolchain}} and \mbox{\path{lake-manifest.json}} pin Lean and
Mathlib. Source locations and related developments are indexed by finding ID.
The offline command \texttt{./reproduce.sh tables} reconstructs the main scores
and mechanism-ablation rates from the released evaluation data.

\subsection{Corpus Census and Visualization}
\label{app:corpus-census}

\paragraph{Development census.} The research collection comprises
22 FOML developments, eight research-paper developments, and three
research-note developments: all 15 benchmark tasks and 18 additional
developments. Its 60 algorithm Lean files contain \NLeanLOC{} physical lines
(approximately 490,000) and 408,470 code-bearing lines. Physical lines include comments and blank
lines; code-bearing lines exclude them. The count follows each algorithm's
local imports, including split files, and excludes shared libraries, external
theory, caches, and unused copies. Algorithm files contain no explicit
\lean{sorry}/\lean{admit} tokens. Counts use the selected paths and
file hashes.
Appendix~\ref{app:source-endpoints} gives the repaired A03 endpoints and
SAPD's theorem scope.

\paragraph{Source availability.}
Sources selected for release cover \NPublicAlgos{} developments, including
MARS: 49 algorithm files, \NPublicLeanLOC{} physical lines, and 339,485
code-bearing lines. The remaining TimeVaryingPushPull development contributes
\NPrivateLeanLOC{} physical lines across 11 files. Its statements and proofs
are temporarily withheld pending publication of the corresponding paper;
the accompanying materials provide an availability placeholder. These lines contribute
to the total corpus size; the source-release count covers the other
\NPublicAlgos{} developments.

\paragraph{Visualization of corpus and reuse.}
Figure~\ref{fig:formalization-scale} shows all 33 developments. Counts exclude
comments and blank lines; reuse includes referenced declarations and their named
dependencies, deduplicated per development. Radial lengths use a square-root
scale, and the inner bars are enlarged by a factor of 1.15 for visibility.
AMSGrad support includes extracted modules used by its proof; shared support is
counted separately from algorithm-local code.

\section{Source Discrepancies and Verification Evidence}
\label{app:source-findings}

\subsection{Counting, Source Versions, and Evidence}

The audit covers \NSourceAuditAlgorithms{} developments: 22 from
\citet{lan2020foml}, eight from research papers, and three from research notes,
matching the development census in Appendix~\ref{app:corpus-census}.
The audit identifies \NSourceIssues{} independent discrepancies affecting
\NSourceIssueAlgorithms{} developments. Each entry identifies its source location,
formalization diagnostic, supporting evidence, and repair scope. Category A contains
\NSourceOriginalIssues{} discrepancies in the selected source versions. Category B
contains \NSourceRevisedIssues{} discrepancies with corresponding author
corrections in later revisions or published errata. A shared mathematical defect is counted once:
VRMD, VRAGD, RAPP, and RGE share A03; 2-RSGD/2-RSGF share A13. Different defects within one
algorithm receive separate IDs.

The catalogue covers incorrect formulas, proof gaps, and algorithm--analysis
mismatches. It includes the specification
and formula corrections A06, A14, A15, A21, A22, and A25. Interpretation-dependent candidates are excluded from this count. Routine
nonzero-denominator conditions, rounding, measurability scaffolding, and absent
library lemmas are tracked outside this count.

Each entry below identifies the source statement, evidence, and the scope of a
repair. The evidence combines formalization diagnostics with independent source review. Lean counterexamples, exact arithmetic checks, and externally
published mathematical counterexamples are distinguished, and each entry states
the scope of its certificate. The evidence identifies source and artifact hashes, declaration locations,
exact arithmetic checks, and the correspondence to these IDs.

Book references use printed page numbers in the 2020 Springer edition.
The selected research versions are SAM v3, PAGE v3, SPIDER v2, STORM v3, Lion v1,
the ICLR 2018 AMSGrad version, PULM-DGD v1, and the ICML 2025 proceedings
version of MARS. Source-version checks include the published errata linked from Lan's publication
page \citep{lan2022errata}. Those corrections address separate defects from
A01--A15. Category B identifies the corresponding author correction explicitly.
Verification of the revised papers lies outside the scope of this catalogue.

\paragraph{How formalization exposes source discrepancies.}
Formalization turns a compressed source step into a Lean obligation with
explicit objects, hypotheses, and dependencies. Audits trace a
blocked obligation back to the corresponding source passage, checking the
encoding and the applicability of the invoked result. They identify the
missing implication or incompatible formula and guide a proof, a corrected
statement, or a counterexample. The entries below combine these diagnostics
with direct checks of the published PDF and the stated mathematical evidence.

For SAM (A17), the reconstruction audit isolated the implication needed to
connect a selected-scale Gaussian argument to the source's radius-scale
premise. Judge flagged the unresolved selected-scale premise when it appeared
as an added theorem hypothesis. Source review and the cross-scale example
in A17 establish the gap. For MARS (A23), applying the estimator-error lemma
required a bridge from the clipped algorithm update to the raw momentum
recursion. The audit identified the condition $\|c_t\|\le1$ needed by that
bridge, and Judge required it to remain explicit in the corrected result.
The published algorithm and proof exhibit the mismatch described in A23.

\paragraph{Research-note targets.}
\textbf{HeavyBall} establishes uniform linear convergence on a specified
parameter box for smooth strongly convex objectives, while excluding
nonconstant finite cycles and a prescribed common two-point Lyapunov certificate.
\textbf{R/C-DGD}, Theorem~4.2, gives an $O(1/T)$ average squared-gradient bound
per communication round and square-summable, vanishing consensus diameters,
under its smoothness, time-varying network, and stepsize assumptions.
\textbf{TimeVaryingPushPull} studies decentralized optimization over
time-varying directed networks. Its detailed statements and Lean proofs
are temporarily withheld pending publication of the corresponding paper.
The release provides an availability placeholder.

\begingroup

\setlength{\tabcolsep}{4pt}
\renewcommand{\arraystretch}{1.1}
\setlength{\LTcapwidth}{\textwidth}
\begin{longtable}{@{}p{.07\linewidth}>{\RaggedRight\arraybackslash}p{.18\linewidth}>{\RaggedRight\arraybackslash}p{.52\linewidth}>{\RaggedRight\arraybackslash}p{.15\linewidth}@{}}
\caption{Complete catalogue of counted source discrepancies. IDs identify independent issues; multiple algorithms sharing one issue appear in the same row. Evidence types describe the scope of the finding, not a uniform claim that final convergence theorems are false.}\label{tab:source-issue-catalog}\\
\toprule
ID & Development & Discrepancy & Evidence scope \\
\midrule
\endfirsthead
\multicolumn{4}{l}{ Table~\thetable{} continued}\\
\toprule
ID & Development & Discrepancy & Evidence scope \\
\midrule
\endhead
\midrule
\multicolumn{4}{r}{ Continued on next page}\\
\endfoot
\bottomrule
\endlastfoot
\multicolumn{4}{@{}l}{\textbf{A. Discrepancies in the selected sources}}\\[2pt]
\hyperref[case:source-A01]{A01} & SNCCG & Corollary 7.12 drops a factor when specializing the stepsize. & Run counterexample \\
\hyperref[case:source-A02]{A02} & SNCCG & The variable-step proof uses the current index for a predecessor update. & Proof step \\
\hyperref[case:source-A03]{A03} & VRMD / VRAGD / RAPP / RGE & A one-sided Bregman bound exceeds the constrained-domain hypotheses. & Lemma counterexample \\
\hyperref[case:source-A04]{A04} & NSAGD & Marginal oracle moments do not justify adaptive conditional centering. & Assumption gap \\
\hyperref[case:source-A05]{A05} & NSBMD & Taking expectation removes a factor $1/2$ from the descent term. & Proof step \\
\hyperref[case:source-A06]{A06} & NSBMD & The printed update substitutes a gradient mapping for the next iterate. & Update formula \\
\hyperref[case:source-A07]{A07} & NVRMD & The descent coefficient loses $\gamma q/2$ before epoch summation. & Proof step \\
\hyperref[case:source-A08]{A08} & RAPP & A reachable component refresh leaves the gradient domain. & Reachable query \\
\hyperref[case:source-A09]{A09} & SAPD & The queried point and the gradient used to center the noise differ. & Oracle mismatch \\
\hyperref[case:source-A10]{A10} & SGS & The final tail exponent is stronger than the displayed concentration bound. & Proof step \\
\hyperref[case:source-A11]{A11} & SGS & The compact quadratic-noise scale is reduced by an invalid inequality. & Proof step \\
\hyperref[case:source-A12]{A12} & VRAGD & The generic parameter conditions allow a negative Jensen weight. & Assumption gap \\
\hyperref[case:source-A13]{A13} & 2-RSGD / 2-RSGF & A fixed-candidate tail bound is reused after data-dependent selection. & Selection counterexample \\
\hyperref[case:source-A14]{A14} & SAMP & Adaptive oracle noise is called independent of its query. & Probability semantics \\
\hyperref[case:source-A15]{A15} & SNCGS & An unweighted average is transferred to an inherited weighted output law. & Output specification \\
\hyperref[case:source-A16]{A16} & SAM & The bounded-loss premise needed by the PAC-Bayes/tail argument is unstated. & Assumption gap \\
\hyperref[case:source-A17]{A17} & SAM & The Gaussian premise is reused at a different variance scale. & Cross-scale counterexample \\
\hyperref[case:source-A18]{A18} & PAGE & The iteration count assumes a larger stepsize than the theorem requires. & Run counterexample \\
\hyperref[case:source-A19]{A19} & Lion & One-sided asymptotic schedules allow steps too small for the claimed rate. & Run counterexample \\
\hyperref[case:source-A20]{A20} & STORM & The final scalar simplification drops a $\sqrt M$ factor. & Proof step \\
\hyperref[case:source-A21]{A21} & SPIDER & The theorem names OPTION I, while its proof uses OPTION II. & Output specification \\
\hyperref[case:source-A22]{A22} & SPIDER & The finite-sum appendix inserts an extra $\epsilon$ in the batch size. & Formula mismatch \\
\hyperref[case:source-A23]{A23} & MARS & The momentum proof omits the clipping used by the algorithm. & Algorithm--proof mismatch \\
\hyperref[case:source-A24]{A24} & MARS & The estimator-error bound loses a factor $\rho$ in its variance term. & Proof step \\
\hyperref[case:source-A25]{A25} & MARS & The simplified variance-reduction parameter loses a momentum factor. & Formula mismatch \\
\addlinespace
\multicolumn{4}{@{}l}{\textbf{B. Discrepancies with subsequent author revisions}}\\[2pt]
\hyperref[case:source-B01]{B01} & AMSGrad & The old momentum-weighted telescope is unsupported by the theorem conditions. & Proof step \\
\hyperref[case:source-B02]{B02} & PULM-DGD & The v1 combined-state update disagrees with the analyzed invariant. & Update / run \\
\hyperref[case:source-B03]{B03} & PULM-DGD & The v1 Lyapunov combination and absorption change coefficients. & Proof step \\
\addlinespace
\end{longtable}
\endgroup

\subsection{Category A: Discrepancies in the Selected Sources}
\label{app:source-category-a}

\subsubsection{Textbook Algorithms}

\paragraph{A01. SNCCG: an incorrect corollary specialization.}
\label{case:source-A01}
Corollary 7.12 (p.~476) specializes Theorem 7.17 with the stepsize in
Eq.~(7.4.15). At zero noise this gives
$\alpha=1/\sqrt{NL\bar D_X^2}$, so the initial-gap term retains a factor
$\sqrt{L\bar D_X^2}$ that the printed specialization omits.
For $X=[0,1]$, $f(x)=100x$, $L=4$, $\sigma=0$, $N=m=4$, $T=b=2$,
and $x_1=1$, the prescribed $\alpha=1/4$ gives
$x_k=(3/4)^{k-1}$. The expected output gap is $4375/64>57$, whereas the
parent bound evaluates to $103$. The existing Lean artifact constructs the
complete instance and its normalized output law. An independent source review
confirms the parameter choices and arithmetic. This refutes the printed
universal corollary, not the parent bound on this instance; the missing-factor
explanation here is specifically for $\sigma=0$.

\paragraph{A02. SNCCG: predecessor and current stepsizes.}
\label{case:source-A02}
The proof of Theorem 7.16 (p.~471), reused for Theorem 7.17 (p.~475), replaces
$\|x_{s,i}-x_{s,i-1}\|^2$ by an expression involving
$\alpha_{s,i}^2\|y_{s,i}-x_{s,i}\|^2$.
The algorithm instead gives
$x_{s,i}-x_{s,i-1}=\alpha_{s,i-1}(y_{s,i-1}-x_{s,i-1})$.
On $X=[0,1]$ with $f(x)=x$, $x_1=1$, $\alpha_1=1$, and $\alpha_2=1/2$,
the linear oracle returns $y_1=y_2=0$ and $x_2=0$: the printed equality has
left side $1$ and right side $0$. Both stepsizes are positive and feasible.
The ensuing maximum must cover predecessor indices, or a schedule comparison
must justify replacing them. This is a local variable-step proof defect;
constant schedules have equal predecessor/current coefficients.

\paragraph{A03. The domain of the one-sided Bregman inequality.}
\label{case:source-A03}
Lan specifies a closed convex feasible set $X$. Lemma 5.8
(pp.~255--256), used in Lemma 5.12 (pp.~280--281), requires smoothness beyond $X$
in both its original proof and the general-norm revision in
\citet[p.~3]{lan2022errata}. The required inequality is
$\|g(x)-g(y)\|^2/(2L)\le f(x)-f(y)-\langle g(y),x-y\rangle$.
The published counterexample of \citet[Section~2]{drori2020open}, restricted to
$[-1,3]\times[-1/20,1]$, gives left side $17545/23040$ and right side
$16991/23040$ at $y=(0,0)$ and $x=(2,0)$. Both points are interior; the
counterexample applies to intrinsic gradients as well.

The agent repairs the VRMD proof by taking $U=\operatorname{ri}(X)$ in the
affine span of $X$ and adding convexity of each intrinsic gradient image $g_i(U)$.
It formalizes open-domain cocoercivity following
\citet[Theorem~3.1]{perezarosvilches}, derives the one-sided bound through
\citet[Lemma~3.3]{wachsmuth2022open}, and extends it to $X$ by continuity.
The resulting Lean proofs connect Lemma 5.12 to Theorem 5.6 and Corollary 5.8
under this explicit additional condition, preserving the coefficient $1/(2L_i)$.
For VRAGD, RAPP, and RGE, the agent instead proves the bound from a whole-space
convex smooth extension with the same function values, selected gradients, and
smoothness constants on $X$. RAPP uses the convex regularized subproblem
$\psi_{i,z}(x)=f_i(x)+\mu\|x-z\|^2$ with constant $L_i+2\mu$.
Table~\ref{tab:source-endpoint-status} summarizes the resulting endpoints and conditions.
The repaired endpoints use proved carrier bridges, with their transitive
proof dependencies checked in Lean.

\paragraph{A04. NSAGD: marginal versus conditional oracle guarantees.}
\label{case:source-A04}
Assumption 16 (p.~361) gives fixed-query marginal mean and variance conditions;
the discussion on p.~374 allows dependent samples. Theorem 6.12's proof
(p.~377) nevertheless uses conditional centering at an adaptive query.
Under the literal marginal interpretation, take
$\Psi(x)=x^2/2$, $G(x,Z)=x+Z$, and reuse one uniform Rademacher variable $Z$
at every iteration. Choose $x_0=0$, $N=5$, $\lambda_k=\beta_k=1/4$,
$\alpha_1=1$, and $\alpha_k=1/2$ thereafter, with independent uniform output.
The two accelerated sequences coincide, with
$x_k=-[1-(3/4)^k]Z$. The left side of Eq.~(6.4.67) is
$69169/327680>1/6$, its printed right side. Independent Lean checks verify
the finite recurrence and marginal moments; this is not a complete
instantiation of the canonical measure-theoretic setup.
Conditional unbiasedness and variance at each adaptive query exclude the
example and supply the required repair. Fresh independent sampling suffices,
but independence itself is not necessary.

\paragraph{A05. NSBMD: a lost factor in expectation.}
\label{case:source-A05}
After Eq.~(6.3.21), the proof on p.~357 has descent coefficient
$(\gamma_k/2)(1-L_{i_k}\gamma_k/2)$. The first expectation inequality on
p.~358 drops the leading $1/2$ without doubling the right side, and is then
used for Theorem 6.9. Taking expectation and averaging the block index do not
justify that change. The corrected source-boundary theorem retains
$(1/2)\mathrm{LHS}\le\mathrm{RHS}$.
The displayed proof therefore supports a bound with twice the printed right
side. This does not establish that the factor two is unavoidable for every
possible analysis of the intended algorithm.

\paragraph{A06. NSBMD: the update uses the wrong mathematical object.}
\label{case:source-A06}
Equation (6.3.8), p.~353, defines $P_X$ as the gradient mapping
$(x-\operatorname{prox}(x))/\gamma$. Algorithm 6.1, Eq.~(6.3.13), p.~354,
then assigns $P_X$ directly as the next block iterate; the subsequent proof
uses the proximal point instead. In one dimension with $X=\mathbb R$,
$\chi=0$, Euclidean prox, $f(x)=x^2/2$, and $\gamma=1/2$, the mapping is
$P_X=x$. Thus the literal update stays at $x_1=1$, rather than moving to
$x-\gamma P_X$. With $N=2$, the printed deterministic bound gives
$1\le2/3$. This elementary check diagnoses a consequential formula/object
error, plausibly a notation typo. It is distinct from A05, which concerns
the proof after using the intended proximal update.

\paragraph{A07. NVRMD: the missing $q$ contribution.}
\label{case:source-A07}
In the proof of Theorem 6.14 (pp.~391--392), combining Eq.~(6.5.10) with the
gradient-mapping estimate drops $\gamma q/2$ from the descent coefficient.
At $\gamma=1/L$, $p=1/(8L)$, and $q=1/8$, the coefficient is
$3/(16L)$, not $1/(4L)$. With the theorem's batch choice $b=17T$,
the printed proof route's residual becomes $(4-T)/(16LT)$, which is
nonpositive for $T\ge4$. The proof text instead uses $b=21T$, yielding
$(68-5T)/(336LT)$, which is nonpositive for integer $T\ge14$.
The canonical development repairs the proof by jointly bounding the
gradient-mapping and estimator-error terms under $\gamma+4p\le2p/q$,
preserving the theorem's original parameters and $b=17T$.

\paragraph{A08. RAPP: a reachable query outside the source domain.}
\label{case:source-A08}
The component functions are defined on $X$ (p.~395), but Algorithm 6.9
(p.~398) extrapolates before refreshing and querying a component.
Take $X=[0,1]$, $m=16$, $L=\mu=1$, and
$f_i(x)=-x^2/2-24x$ for every component. Initialize all primal memories at
zero and all component gradients at $-24$. Theorem 6.16's parameters give
$\alpha=23/24$, $\tau=1/2$, and $\eta=23$.
The first refreshed point remains zero; the first proximal objective is
$12x^2-24x$, uniquely minimized on $X$ at $x^1=1$.
At the second step every possible sampled component is refreshed to
$[(23/24)(1-0)+1]/(1+1/2)=47/36>1$.
Thus the domain violation is reachable, not an arbitrary combination of
states. Finite Lean certificates check the proximal minimizer and this
arithmetic. A repair must extend the function/oracle assumptions to the
required query domain or establish a different invariant. The claim concerns
well-definedness of the printed algorithm, not failure of an extended algorithm.

\paragraph{A09. SAPD: the oracle query and noise center disagree.}
\label{case:source-A09}
Both the published book and the benchmark source package specify
$\widehat G(x_t)$ in Algorithm 4.3, Eq.~(4.4.57), p.~170;
the proof on pp.~171--175 centers that quantity at
$\nabla\widehat f(x_t^{md})$ and applies the same-query variance assumption.
For the exact oracle $G(x)=x$ of $f(x)=x^2/2$, the same-query noise has
variance zero, whereas the mixed-center squared error is one at
$x_t=1$, $x_t^{md}=0$. This gives a deterministic counterexample to the variance implication. Querying at $x_t^{md}$, or carrying
the deterministic gradient difference in the proof, repairs the mismatch.
The evaluated artifact follows the printed query, identifies this discrepancy,
and proves bounds under explicit additional premises. Its theorem coverage is
summarized in Table~\ref{tab:source-endpoint-status}. A separate repair artifact
proves a corrected expected-gap theorem with an explicit gradient-mismatch
budget, under stated same-sample independence and martingale-cancellation
premises. Its high-probability counterpart remains a proof target; the benchmark
ratings refer to the evaluated fixed version.

\paragraph{A10. SGS: an unsupported tail exponent.}
\label{case:source-A10}
Theorem 8.2(b), Eq.~(8.1.67), p.~499, claims
$\exp(-2\lambda^2/3)+\exp(-\lambda)$.
The displayed martingale bound (8.1.70), p.~501, supplies only
$\exp(-\lambda^2/3)$, and (8.1.71) supplies $\exp(-\lambda)$.
Their union does not yield the stronger printed exponent.
The repair is to retain the supported exponent or supply an independently
sharper concentration proof. This is a failure of the displayed derivation,
not a complete stochastic counterexample to every possible proof of the theorem.

\paragraph{A11. SGS: the compact quadratic-noise scale.}
\label{case:source-A11}
For Corollary 8.3(b), the compact schedule in Eq.~(8.1.42) includes
$1-P_{T_k}$ in $\beta_k$. Substitution into the quadratic part of
Eq.~(8.1.68) therefore gives the normalized row
\[
 r(T)=\frac{P_T}{(1-P_T)^2}
       \sum_{i=1}^T\frac{1}{p_i^2P_{i-1}},\qquad
 p_i=i/2,\quad P_i=\frac{2}{(i+1)(i+2)}.
\]
The reduction on p.~503 would require $r(T)\le4/T$, but
$r(9)=327899/734832>4/9$. This value is source-admissible:
$N=L=\widetilde D=\sigma^2=1$, $M=1/100$ gives $T_1=9$.
The inequality is independently checked by exact rational arithmetic and
verified in Lean. Keeping the full quadratic scale gives a
safe term $16(1+\lambda)\widetilde D/3$ along the checked route, instead of
$8(2+\lambda)\widetilde D/3$. This is a local scale-compression error;
neither optimality of that repair nor falsity of the full corollary is asserted.

\paragraph{A12. VRAGD: a negative convex-combination weight.}
\label{case:source-A12}
Lemma 5.16 (pp.~289--290) uses Jensen's inequality with weights
$(1-\alpha-p,\alpha,p)$ without ensuring $\alpha+p\le1$.
Take $m=q_1=L_1=1$, so $L=L_Q=1$, and choose
$\mu=0$, $\gamma=1/4$, $\alpha=p=3/4$.
The quantities constrained by (5.4.7)--(5.4.8) have positive margins
$13/16$ and $27/52$, yet the first weight is $-1/2$.
The finite Lean check satisfies the required relation $L\le L_Q$.
Adding the nonnegative-weight condition repairs this generic lemma interface.
The prescribed schedule of the final theorem can satisfy it, so this example
does not refute that specialized convergence result.

\paragraph{A13. Two-stage methods: concentration after selection.}
\label{case:source-A13}
For 2-RSGD, Eqs.~(6.1.39) and (6.1.45), pp.~317--318, apply a
fixed-candidate tail estimate after selecting the candidate with the smallest
estimated gradient norm using the same validation data.
Let two candidates have true gradient one and one validation sample each.
Independent errors take values $-1$ and $1/3$ with probabilities $1/4$ and
$3/4$. Every fixed candidate is unbiased with variance $1/3$.
At $\lambda=5/2$, however, the selected error exceeds the squared-error
threshold $5/6$ with probability $7/16>2/5=1/\lambda$.
This four-atom calculation is exact. It can be realized by a uniform
fixed-query oracle for $f(x)=\sin x$ at candidates $0$ and $2\pi$;
we do not claim to realize their complete preceding optimization phase.
The same unsupported inference occurs in 2-RSGF's Eq.~(6.1.90), p.~327.

The final guarantee can be preserved. With
$m=\min_s\|\nabla f(x_s)\|^2$ and
$M=\max_s\|\widehat g_s-\nabla f(x_s)\|^2$, Eq.~(6.1.36) directly gives
$\|\nabla f(x_{s^*})\|^2\le4m+6M$.
The failure probability is then bounded by $2^{-S}+S/\lambda$, which is
stronger than the printed $2^{-S}+(S+1)/\lambda$.
The light-tail argument admits the analogous maximum-error repair.
The 2-RSGF formalization implements this route; its Gaussian
estimator is not claimed to instantiate the four-atom example above.
This is one shared proof defect, not two false convergence theorems.

\paragraph{A14. SAMP: independence is stronger than the needed centering.}
\label{case:source-A14}
The proof on p.~191 describes the evaluated adaptive noise as independent of
its query. This need not hold even with fresh independent oracle samples.
For example, let $H(u,\zeta)=u\zeta$ with independent Rademacher samples.
An initial half-step can give $w_2=1/4$ or $3/4$, and the next noise is
$\Delta H_2=w_2\zeta_2$. Its absolute value determines $w_2$, so the two
quantities are not independent, although
$\mathbb E[\Delta H_2\mid\text{past}]=0$.
Conditioning on the interleaved history supplies the cancellation actually
needed. We record a probability-semantics correction, not a counterexample
to the final convergence guarantee. The source already states unbiasedness;
we do not count an additional missing-unbiasedness issue.

\paragraph{A15. SNCGS: the output law must match the average proved.}
\label{case:source-A15}
Theorem 7.18 (pp.~479--481) proves an unweighted sum of squared gradient
mappings and then invokes the output distribution. The algorithm modification
in Eq.~(7.5.5) replaces the iterate update, while the referenced Algorithm 7.13
uses probabilities proportional to $\alpha_k$.
Without uniform sampling, constant weights, or a weighted descent argument,
an unweighted mean does not establish that output expectation: values $(1,0)$
with weights $(0.9,0.1)$ have weighted mean $0.9$ but uniform mean $0.5$.
The existing development proves the uniform-output version. This diagnoses
an output specification gap; the abstract two-value example is not a
complete counterexample trajectory of SNCGS.

\subsubsection{Research-Paper Algorithms}

\paragraph{A16. SAM: the loss range required by the proof.}
\label{case:source-A16}
The setup in \citet{foret2021sam} allows nonnegative losses, whereas the
bounded-loss PAC-Bayes inequality invoked in Appendix A.1 and the Gaussian
tail contribution in the proof of Theorem 2 require a bounded range.
In particular, bounding the outside-neighborhood loss contribution by its
probability uses an upper bound of one on the loss; nonnegativity alone
does not supply it. A constant loss of two already defeats that particular
tail estimate, without refuting the final sharpness inequality.
The checked formulation states $0\le\ell\le1$ explicitly.
For other loss classes, the proof needs a range-dependent or tail-sensitive
PAC-Bayes bound and a corresponding outside-neighborhood term.
This is a substantive assumption gap, not a claim that SAM's optimization
algorithm is invalid.

\paragraph{A17. SAM: the Gaussian scale in the premise changes.}
\label{case:source-A17}
Theorem 2 assumes that Gaussian averaging at scale $\rho$ does not decrease
the population loss. The proof selects the smaller scale
\[
 \sigma=\frac{\rho}{\sqrt{k}(1+\sqrt{\log(n)/k})}
\]
to obtain the required radius concentration, and then uses the same
nondecrease premise at $\sigma$. The cross-scale implication need not hold,
even for a smooth bounded loss. For
$\ell(x)=0.5-0.2\cos x+0.1\cos(2x)$, we have
$0.2\le\ell\le0.8$ and $\ell(0)=0.4$.
At $k=1$, $n=16$, and $\rho=2$, the Gaussian average is approximately
$0.473$ at $\rho$ but $0.382$ at the selected $\sigma$.
These values follow from the Gaussian characteristic function.
This independent mathematical example strengthens an abstract profile
obstruction in the Lean record; it is not a complete Lean counterexample
to the final generalization theorem.
The repair is to state the premise at the selected scale, or uniformly over
the relevant scales, and use that same scale throughout the proof.

\paragraph{A18. PAGE: the iteration count requires a specified stepsize.}
\label{case:source-A18}
Theorem 1 of \citet{li2021page} permits any positive
$\eta\le\eta_{\max}$ but prints the horizon
$T=2\Delta_0/(\epsilon^2\eta_{\max})$.
Appendix Eq.~(24) identifies it with $2\Delta_0/(\epsilon^2\eta)$,
which requires equality of the stepsizes.
Take two identical components $f_i(x)=x^2/2$, $x_0=1$, $L=1$, $p=1$,
$\eta=1/1000$, and $\epsilon=1/10$. The printed horizon is $T=100$,
but $x_t=(999/1000)^t\ge1-t/1000\ge901/1000$ for $0\le t<100$.
Thus the uniform output's expected gradient norm exceeds $0.9$, not $0.1$.
This is an elementary actual-run counterexample to the literal guarantee;
the existing Lean record also identifies the failed scalar specialization.
The repair is to require $\eta=\eta_{\max}$ for that horizon, or let the
horizon depend on the actual $1/\eta$.

\paragraph{A19. Lion: one-sided asymptotic schedules are insufficient.}
\label{case:source-A19}
Theorem 1 of \citet{jiang2025lion} specifies
$\beta_2=O(T^{-1/2})$ and $\eta=O(d^{-1/2}T^{-3/4})$, while its proof
requires reciprocal-rate control as well.
In dimension one, use the deterministic objective $f(x)=(x-1)^2/2$,
$x_1=0$, zero weight decay, and positive parameters
$\beta_1=\beta_2=\eta=(T+1)^{-2}$.
They satisfy the printed upper-order conditions, but total movement is at
most $T/(T+1)^2$, so the average true-gradient norm stays at least $1/2$
instead of decaying as $O(T^{-1/4})$.
The canonical Lean development contains a positive-parameter quadratic
counterexample and a corrected endpoint with reciprocal-rate conditions.
Explicit parameter schedules, or appropriate two-sided rates, repair the
statement. This targets the literal quantifiers; it does not refute Lion
with the intended tuning, and the printed notation may reflect informal
same-order shorthand. Distributed and compressed variants are outside this entry.

\paragraph{A20. STORM: a missing factor in the final simplification.}
\label{case:source-A20}
In the selected v3 of \citet{cutkosky2019momentum}, the last scalar split
in the proof of Theorem 1 (p.~8) removes the $\sqrt M$ dependence from the
stochastic contribution. Splitting
$\sqrt{2M}(w+2T\sigma^2)^{1/6}/\sqrt T$ gives the valid stochastic term
$2^{2/3}\sqrt M\,\sigma^{1/3}/T^{1/3}$, rather than
$2\sigma^{1/3}/T^{1/3}$ without an additional restriction on $M$.
Here $M$ includes the initial objective gap and a logarithmic horizon term.
The formalization exposes a scalar obstruction and retains the valid
preceding bound
\[
 \frac{\sqrt{2M}(w+2T\sigma^2)^{1/6}+2M^{3/4}}{\sqrt T}.
\]
This changes explicit parameter dependence, while the displayed powers of
$T$ and $\sigma$ are unchanged. It is a local proof error, not a complete
stochastic-run counterexample to the final theorem.

\paragraph{A21. SPIDER: the named option differs from the option proved.}
\label{case:source-A21}
Theorem 1 in the selected v2 of \citet{fang2018spider} names OPTION I.
Its proof, however, uses the clipped adaptive stepsizes and uniform output
of OPTION II. OPTION I has a stopping test and a different output rule.
The formalization exposes the mismatch explicitly and keeps the
expectation theorem attached to its actual output convention.
Relabeling that result as OPTION II repairs this correspondence; an OPTION I
guarantee requires an argument for the stopped process.
We do not infer that the intended first-order convergence rate is false.

\paragraph{A22. SPIDER: the finite-sum batch-size formula changes.}
\label{case:source-A22}
The finite-sum schedule in Eq.~(3.7) sets $S_2=\sqrt n/n_0$.
Appendix B.2 instead substitutes $S_2=\sqrt n/(\epsilon n_0)$
in the calculation leading to (B.18), while retaining the printed
$\epsilon^2$ result; the substituted product has an $\epsilon^3$ factor.
The source constants and the appendix calculation therefore cannot both
be used literally. Restoring the finite-sum batch-size formula repairs this
step. The Lean development contains an exact scalar check of the
factor mismatch. This is a formula correction, separate from the output
specification in A21, and not a claim that integer rounding is an error.

\paragraph{A23. MARS: clipping is omitted from the momentum recursion.}
\label{case:source-A23}
Algorithm 1 of \citet{yuan2025mars} (p.~4, lines 5--6) updates momentum
with $\widetilde c_t=\operatorname{Clip}(c_t,1)$. The proof of Lemma C.2
in Section D.4 (p.~23) instead expands the update using $c_t$ directly.
Writing $\beta=\beta_{1,t+1}$ and
$r_{t+1}=\operatorname{Clip}(c_{t+1},1)-c_{t+1}$, the actual update is
\[
 m_{t+1}=\beta m_t+(1-\beta)c_{t+1}+(1-\beta)r_{t+1}.
\]
The displayed proof omits the last term. Assumptions B.1--B.3 give bounded
oracle variance, smoothness, and a lower bound on the preconditioner;
these permit corrections with norm above one. For example, in one dimension,
$m_t=0$, $c_{t+1}=2$, and $\beta=1/2$ give actual momentum $1/2$, whereas
the expanded recurrence gives $1$. This checks the update identity locally.
The Lean repair makes $\|c_t\|\le1$ explicit when using the unclipped
recursion. An analysis covering active clipping must retain and bound
$r_{t+1}$ in the estimator-error argument.

\paragraph{A24. MARS: a lost preconditioner factor in the variance term.}
\label{case:source-A24}
Equation (C.8), p.~19, weights the estimator error by $\eta_t/\rho$
and has variance contribution $\rho c^2\sigma^2\log(s+T)/(8L^2T)$.
Extracting the unweighted estimator-error bound multiplies this term by
$\rho/\eta_T$, producing
$\rho^2c^2\sigma^2\log(s+T)/(8L^2T\eta_T)$.
The next two displays retain $\rho^2$, but the final simplification and
Theorem B.5 (p.~16) use $\rho c^2\sigma^2/(4L^2T^{2/3})$ as the
logarithmic coefficient. Assumption B.3 allows every $\rho>0$; the stated
condition $T\ge s$ leaves this additional factor uncontrolled.
The same coefficient change recurs in the proof of Theorem B.6 (p.~21)
and is counted once. A Lean scalar identity checks the extra $\rho$
introduced by extraction. Retaining $\rho^2$ repairs this coefficient
step; the algorithm--proof correspondence is addressed separately in A23.

\paragraph{A25. MARS: the parameter simplification drops a momentum factor.}
\label{case:source-A25}
In Eq.~(D.12), p.~24, let $\beta=\beta_{1,t+1}$,
$V=\mathbb E\|\Delta_t\|^2$, $u=\|\mathbb E\Delta_t\|^2$, and
$G=G_{t+1}$. For $\beta V\ne0$, the first expression simplifies to
\[
 1-\frac{G+\beta(V-u)}{\beta V}
 =\frac{\beta u-G}{\beta V},
\]
whereas the printed final numerator is $u-G$.
At $\beta=1/2$, $V=1$, $u=1/4$, and $G=0$, exact arithmetic gives
$1/4$ for the first expression and $1/2$ for the printed simplification.
Restoring $\beta u$ preserves the parameter choice derived from the
preceding quadratic minimization. This certificate concerns the scalar
identity used to select the variance-reduction parameter.

\subsection{Category B: Discrepancies with Subsequent Author Revisions}
\label{app:source-category-b}

These findings remain part of the total because the selected input versions
contain the documented defects. Later author revisions are identified explicitly;
the findings concern the selected versions and do not assert that revised papers
retain the same defects.

\paragraph{B01. AMSGrad: the old momentum-weighted telescope.}
\label{case:source-B01}
In the selected ICLR 2018 version of \citet{reddi2018convergence},
Theorem 4 gives $\beta_{1,t}\le\beta_1$; the surrounding discussion
mentions decreasing momentum. The theorem's displayed conditions alone
do not support the signed telescoping comparison in Appendix D, Eq.~(18).
For $\beta_{1,t}=(0.9,0,0.9)$, $\alpha_t=1/\sqrt t$,
$\sqrt{\widehat v_t}=1$, and squared distances $(0,0,1,0)$, the original
expression exceeds its proposed upper bound by $9\sqrt2/2$.
An independent Lean certificate checks this scalar inequality; it does not
construct a complete AMSGrad run or refute every possible regret proof.
Nonincreasing $\beta_{1,t}$ supports a repair of the original telescope,
and the selected formal development exposes that condition at its endpoints.

The later author revision \citep{reddi2019revision} changes
the estimate and explicitly acknowledges a missing factor in footnote 3.
The second coefficient in Theorem 4 changes from
$D_\infty^2/[2(1-\beta_1)]$ to $D_\infty^2/(1-\beta_1)^2$.
That revised theorem does not add monotonicity.
Thus this is a known, version-specific proof gap; we do not claim that
monotonicity is necessary for every AMSGrad analysis or that the revised
paper still has this defect.

\paragraph{B02. PULM-DGD: the v1 update and the analyzed invariant.}
\label{case:source-B02}
Algorithm 3 in the selected v1 of \citet{liang2025pulm} initializes
$z=x-\gamma g$ and updates $z^+=Az-Dg$.
The gradient-averaging recursion is $W^+=AW-D$, so the intended invariant
$z=A_{\mathrm{prod}}x-\gamma Wg$ instead requires
$z^+=Az+\gamma Dg$.
The literal v1 update gives
\[
 z_R=A_{\mathrm{prod}}x-
       [(\gamma+1)A_{\mathrm{prod}}-W_R]g,
\]
as also established by the exact matrix recurrence in the canonical artifact.
This changes the effective gradient coefficients, not just a bound constant.

An independently derived deterministic example uses two nodes,
$A=\left(\begin{smallmatrix}3/4&1/4\\3/4&1/4\end{smallmatrix}\right)$,
$f_1(x)=(x-1)^2/2$, $f_2(x)=(x+1)^2/2$, consensus $x_0=0$, and
$\gamma=1/100$. The memory matrices satisfy
$\|W_R-J\|_2\le(3/4)^R$, with $J$ the averaging matrix.
Using 100 inner rounds for the first $10^4$ outer iterations, the actual
v1 process is within $10^{-8}$ of the consensus trajectory
$y_k=50.5(1-0.99^k)$.
Its average squared global gradient exceeds $618.75$, while the printed
bound is $0.09$. The archived derivation checks the stepsize and mixing
conditions and consumes each time-indexed communication instance once;
equal numerical matrices are not replayed communication events.
The full trajectory counterexample is established by this mathematical
derivation; the Lean artifact checks the matrix recurrence.
Version v3 \citep{liang2026pulmrevision} uses the required $+\gamma Dg$ update.
The v1 counterexample is not attributed to v3.

\paragraph{B03. PULM-DGD: v1 coefficient combination and absorption.}
\label{case:source-B03}
Appendix D of v1, pp.~30--31, changes the objective-gap coefficient
$40n\gamma L C_M^2$ to a Lemma 6 application with
$c=8nC_M^2L$; the consensus coefficient also changes from
$12/\gamma$ to $8/(n\gamma)$.
Moreover, multiplying Eq.~(18) by the stated $24/\gamma$ already
contributes $192n\gamma L C_M^2$ before the descent contribution.
The displayed combination therefore does not justify the printed
absorption constants and associated admissible stepsize.
The canonical artifact confirms the incompatible scalar coefficients.
Author v3 rewrites the theorem's stepsize and communication-round conditions
alongside the algorithm revision. We retain this as a separate v1 proof
defect because repairing the update alone does not supply the missing
coefficient derivation; we do not claim to have verified every constant
in the revised proof.

\subsection{Repaired Endpoints and Evaluation Scope}
\label{app:source-endpoints}

Table~\ref{tab:source-endpoint-status} summarizes the A03 repairs and
the SAPD assessment. The A03 checks cover 17 named algorithm declarations,
including supporting lemmas and the convergence endpoints listed below.
Their proofs and transitive dependencies are checked in Lean. The endpoint checks include declaration names, signatures, and per-endpoint dependency checks.
The added domain conditions appear explicitly in each algorithm's setup;
the other premises of each named endpoint remain part of its statement.

\begin{table}[!htbp]
\caption{Endpoint status for A03 and SAPD. Extension conditions preserve
function values and selected gradients on $X$. A03 repairs add explicit
mathematical premises to the selected endpoint statements.}
\label{tab:source-endpoint-status}
\centering
\renewcommand{\arraystretch}{1.15}
\setlength{\tabcolsep}{4pt}
\begin{tabular}{@{}>{\RaggedRight\arraybackslash}p{.14\linewidth}>{\RaggedRight\arraybackslash}p{.28\linewidth}>{\RaggedRight\arraybackslash}p{.53\linewidth}@{}}
\toprule
Development & Endpoint scope & Repair condition and proof status \\
\midrule
VRMD & Theorem 5.6; Corollary 5.8 and its complexity bound &
Each $g_i(\operatorname{ri}(X))$ is convex. The repaired bounds are proved. \\
VRAGD & Theorem 5.9; corrected-core Corollary 5.10; scalar-handoff Corollary 5.11 &
Each component admits a convex smooth extension preserving $L_i$.
These endpoint variants are proved under their stated premises. \\
RAPP & Source-domain-corrected Theorem 6.17; outer-prefix variant of Theorem 6.16 &
Each regularized subproblem admits a convex smooth extension preserving
$L_i+2\mu$. The variants retain their source-domain and initialization premises. \\
RGE & Canonical-policy Theorem 5.4 &
Components with $L_i>0$ admit convex smooth extensions preserving the
original norm and $L_i$. The endpoint is proved. \\
SAPD (evaluated) & Theorem 4.8(a,b), source-level claims &
The artifact identifies the oracle-query mismatch and proves bounds under
explicit premises. The full expected-gap and tail claims remain proof targets;
the light-tail condition is represented by an opaque predicate. \\
\bottomrule
\end{tabular}

\end{table}

\FloatBarrier

\section{System Implementation and Worked Example}
\label{app:implementation}

The supplied engine implements proof construction, revision, and certification
through separate agents, contracts, and Lean inspection tools. This appendix
describes their interaction, the audit checks, and the prompt instructions that
govern their work, followed by a worked SAM example.
Task-level execution settings are specified in Appendix~\ref{app:evaluation};
library retrieval and write-back are detailed in
Appendix~\ref{app:soptlib-experiments}.

\begin{algorithm}[!t]
\caption{Planner-routed formalization control loop}
\label{alg:full-pipeline}

\begin{algorithmic}[1]
\Require source $S=(A,T,P)$; initial knowledge base $K_0$
\Ensure certified endpoint; updated knowledge base $K$
\State $(\widehat T,X) \gets \Call{Model}{S,K_0}$ \Comment{$X=(L,H)$}
\While{\textbf{true}}
  \State $(L,H) \gets X$; $(E,\Sigma^{\mathrm{lock}}) \gets
    (\Call{Inspect}{L},\operatorname{Lock}(\operatorname{Protected}(L,H)))$
  \If{$\Call{MechanicallyClosed}{E}$}
    \State $V \gets \Call{Certify}{S,L,H}$; \textbf{if} $\Call{Released}{V}$ \textbf{ then break};
      $H \gets H \mathbin{\oplus} \Call{Record}{V}$
  \EndIf
  \State $R \gets \Call{Planner}{S,\widehat T,K_0,L,E,H,\Sigma^{\mathrm{lock}}}$
  \If{$R=\textsc{ProverStep}(B)$}
    \State $(L',\tau) \gets \Call{Prover}{B,L,H}$
    \State $a \gets \Call{Audit}{S,B,L',\tau}$; $H \gets H \mathbin{\oplus} (B,\tau,a)$
    \State \textbf{if }$\operatorname{AdmitProve}(L,L',a,\Sigma^{\mathrm{lock}})$\textbf{ then } $L \gets L'$
  \Else \Comment{$R=\textsc{Blocker}(\beta)$}
    \Repeat
      \State $(C,L',\tau) \gets
        \Call{Refactor}{\beta,S,\widehat T,L,H,\Sigma^{\mathrm{lock}}}$
      \State $E' \gets \Call{Inspect}{L'}$;
        $J \gets \Call{Judge}{C,S,L',E',H}$
      \State $H \gets H \mathbin{\oplus} (C,\tau,E',J)$;
        $(L,\widehat T) \gets (L',\operatorname{Realization}(L'))$
    \Until{$\operatorname{ReviewedRevision}(J,E') \lor \operatorname{EndRound}(J,H)$}
  \EndIf
  \State $X \gets (L,H)$
\EndWhile
\State $K \gets \Call{WriteBack}{L,H,V,K_0}$ \Comment{Extract, generalize, merge, rebuild}
\State \Return $(L,V,K)$
\end{algorithmic}
\end{algorithm}

\subsection{Formalization Control Loop}
\label{app:control-loop}

Algorithm~\ref{alg:full-pipeline} gives the execution loop for $X=(L,H)$, where
$L$ is the Lean development and $H$ is the proof state and audit context.
Inspect returns $E$, containing goals, diagnostics, declarations, and dependencies.
Planner returns a proof blueprint $\textsc{ProverStep}(B)$ or a localized
$\textsc{Blocker}(\beta)$ for contract revision. Here $\widehat T$ denotes the current
$T_t^\star$, $\Sigma^{\mathrm{lock}}$ is the protected-interface state, and $V$ is
the certification verdict. Each candidate produces execution evidence $\tau$ and an Audit verdict $a$.

$\operatorname{AdmitProve}$ checks Lean acceptance, Audit's source correspondence and
proof-route verdict, and an empty protected-signature diff
$\operatorname{PDiff}(L,L';\Sigma)$. This diff covers changed or removed protected
heads and new declarations with protected roles. In reconstruction, Judge examines
the edited candidate, its contract, and Lean evidence. $\operatorname{ReviewedRevision}$
requires a confirmed review, successful boundary checks, and resolution of the
original reconstruction trigger. $\operatorname{EndRound}$ returns a budget limit
or a handoff to another task; such an exit carries the remaining obligations.

The controller carries proof states, contracts, Lean outcomes, and reviews
between reconstruction and proof construction. Remaining obligations accompany
any handoff to proof construction. After all Lean obligations close,
the development is cleaned up and frozen. Certify checks the completed development;
success produces a release certificate, while failure returns diagnostics to Planner.

\paragraph{Proof-state inspection and progress review.}
Planner inspects the current Lean goals and hypotheses before selecting a proof
unit. If a source-facing root remains unresolved across rounds, a separate
strategy audit traces the dependency chain from the active helper to that root.
It examines whether new lemmas are consumed downstream and whether the route
matches the source proof, then supplies a concrete continuation or rerouting
recommendation. The audit therefore assesses mathematical progress alongside
changes in the number of open obligations.

\subsection{Faithfulness Audit Implementation}
\label{app:faithfulness-audit}
\label{sec:appendix}

This subsection details the mechanisms summarized in Section~\ref{sec:faithful}.
Appendix~\ref{app:signature-contract-workflow} describes the contract-based revision loop.

\subsubsection{Protected Interfaces and Change Tracking}

\paragraph{Protected signatures and semantic checks.}
On every \textsf{prove} edit, the anti-tamper check compares declaration heads
and \textsc{Setup}-field types with the locked source-facing signatures.
Deleting a protected head reverts the file; changing one restores its signature
while retaining the proof body. Audit checks the resulting definitions and proof
dependencies against the source argument. During \textsf{reconstruct}, Judge
examines changes to definition bodies, instances, and notation using the
signature contract and source evidence. Certify checks source correspondence
and the full dependency closure at the endpoint.

\paragraph{Signature-level diff.}
Each \textsf{reconstruct} edit produces a hash-based diff of source-facing heads
and \textsc{Setup}-field types. The diff associates changes with the blocker's
proof map and flags undisclosed edits. Judge reviews it together with
the actual definitions and the contract's evidence for their source correspondence.

\subsubsection{Revision Contracts and Judge Review}
\label{app:signature-contract-workflow}

\paragraph{Trigger and proposal.}
Ordinary proof steps preserve the protected signatures. When Planner identifies
an interface or modeling mismatch that blocks a proof, it routes the obligation
to \textsf{reconstruct}. Refactor proposes a revision through a signature
contract: a record of the affected declarations, proposed operation, source
evidence, and dependent proofs. The \texttt{source\_role} field distinguishes
source assumptions from facts derived in the proof; \texttt{book\_citation}
identifies the supporting passage.

\paragraph{Review and revision.}
Judge compares the contract with the actual changes and the cited source.
An added premise must be supported as an assumption of the source result;
a fact derived by the source remains a proof obligation. For a changed
representation, Judge checks that its definitions and properties realize
the source object and preserve the target claim. Feedback identifies the
unsupported change or missing justification, and Refactor revises the
candidate and its contract. A passing decision receives a second independent
Judge review; disagreement returns the candidate for revision.

\paragraph{Proof continuation and validation.}
The reviewed interfaces determine the next proof obligations. These include
lemmas introduced by the revision, representation bridges, and affected
downstream proofs. Lean checks the resulting declarations and proofs, and
Certify checks the completed dependency chain against the source. The
contract thus connects a change in the model to both its source justification
and the mathematical work required to complete the proof.

\paragraph{Example: SAM's Gaussian KL bridge.}
The SAM proof needs to apply a Gaussian KL formula across two representations
of the same probability laws. The initial bridge allowed an arbitrary
measurable structure, while the coordinate transport required the Borel
structure of the source's Euclidean parameter space. Planner therefore
routed the blocker to reconstruction. The proposed revision makes that
structure explicit on the bridge and its consumers. In contract terms,
the source's parameter domain justifies the representation change, while
the KL identity remains a lemma to prove. Judge accepted the Euclidean
realization; the construction establishes the measurable coordinate map
and proves KL transport through it. The downstream PAC-Bayes proof then
uses the proved bridge. Appendix~\ref{app:sam-construction} gives the
mathematical construction.

\subsubsection{Endpoint Guarantees and Remaining Obligations}

\paragraph{Conservativity at the endpoint.}
\textsf{prove} preserves protected signatures, while \textsf{reconstruct}
submits proposed changes for review. At a kernel-certified endpoint,
the internal fragment $D_{\mathrm{int}}$ contains definitions and proved lemmas
with checked dependencies. These extensions preserve the source assumptions.

\paragraph{Remaining obligations.}
Reconstruction exposes unresolved lemmas and the evidence supporting their next
proof attempts. Compiler-grounded feasibility evidence and proposed external results
guide this work; final certification still requires completed Lean proofs throughout
the target's dependency closure.

\paragraph{Evidence bound to the reviewed files.}
Content hashes bind the source, target files, entry state, review artifacts,
release roots, and mechanical and semantic verdicts. Validation checks these
bindings when accepting a candidate, so later edits can be distinguished from
the reviewed version.

\subsection{Agent Roles and Prompt Instructions}
\label{app:agent-prompts}

Figure~\ref{fig:agent-topology} shows the roles and review loops in the supplied
engine. A Source producer supplies the source-grounded starting package. The Model
panel then includes object-model refactoring, a modeling audit, and an alignment
judge before the protected interface is handed to proof construction.
Planner and reviewers inspect the source, live Lean state, and prior attempts;
editing agents receive a localized goal and the applicable signature constraints.
Review feedback determines the next proof step or the scoped reconstruction.

\begin{figure}[!htbp]
\centering
\begin{tikzpicture}[
  x=1cm,y=1cm,
  font=\fontfamily{lmss}\selectfont\fontsize{9}{10.5}\selectfont,
  role/.style={draw=reaslab!48!white,line width=.55pt,
    rounded corners=3pt,fill=white,align=center,
    text width=3.1cm,minimum height=.70cm,inner sep=4pt},
  review/.style={role,draw=rej!42!white,fill=rej!5!white},
  memory/.style={role,draw=reaslab!60!white,fill=boxbg},
  flow/.style={-{Stealth[length=1.8mm,width=1.25mm]},
    line width=.65pt,draw=reaslab!85!black,rounded corners=2pt},
  feedback/.style={flow,dash pattern=on 2.4pt off 1.8pt,
    draw=black!48},
  edge/.style={font=\fontfamily{lmss}\selectfont\fontsize{8}{9.5}\selectfont,
    text=black!70,fill=white,rounded corners=1.5pt,inner xsep=3pt,inner ysep=1.5pt},
  stage/.style={font=\fontfamily{lmss}\selectfont\bfseries\fontsize{10}{12}\selectfont,
    text=reaslab,anchor=west},
  panel/.style={draw=reaslab!18!white,line width=.5pt,
    rounded corners=5pt,fill=reaslab!2!white}
]
\begin{scope}[on background layer]
  \draw[panel] (0,.45) rectangle (14.45,-1.50);
  \draw[panel] (0,-2.0) rectangle (14.45,-5.65);
  \draw[panel] (0,-6.35) rectangle (14.45,-9.55);
  \draw[panel] (0,-10.10) rectangle (14.45,-14.45);
\end{scope}

\node[role,fill=boxbg,text width=3.5cm] (sourceproducer) at (7.25,1.25)
  {Source producer};
\node[stage] at (.35,.15) {Model};
\node[role] (model) at (2.60,-.65) {Object-model refactor};
\node[review] (modelaudit) at (7.15,-.65) {Model audit};
\node[review] (alignment) at (11.70,-.65) {Alignment judge};
\draw[flow] (sourceproducer.south) -- (7.25,.72) -- (3.10,.72)
  -- ([xshift=5mm]model.north);
\draw[flow] (model.east) -- (modelaudit.west);
\draw[flow] (modelaudit.east) -- (alignment.west);
\draw[feedback] (modelaudit.south) -- (7.15,-1.28)
  -- node[edge,above] {repair} (2.60,-1.28) -- (model.south);
\draw[feedback] (alignment.south) -- (11.70,-1.28)
  -- (2.60,-1.28) -- (model.south);

\node[stage] at (.35,-2.22) {Reconstruct / Refactor};
\node[role,text width=2.7cm] (planner) at (2.60,-3.20) {Planner};
\node[role,text width=2.7cm] (prover) at (7.15,-3.20) {Prover};
\node[review,text width=2.7cm] (audit) at (11.70,-3.20) {Audit};
\node[role,text width=2.5cm] (refactor) at (3.40,-4.65) {Refactor};
\node[review,text width=2.5cm] (judge) at (10.60,-4.65) {Judge};
\draw[flow] (alignment.east) -- (14.10,-.65) -- (14.10,-1.74)
  -- (-.25,-1.74) -- (-.25,-3.20) -- (planner.west);
\draw[flow] (planner.east) -- node[edge,above=2pt] {plan} (prover.west);
\draw[flow] (prover.east) -- (audit.west);
\draw[feedback] (audit.south) -- (11.70,-3.97)
  -- node[edge,above=2pt] {route review} (2.60,-3.97) -- (planner.south);
\draw[flow] (planner.south) -- (2.60,-3.62) -- (.45,-3.62)
  -- (.45,-4.65) -- node[edge,above=2pt] {blocker} (refactor.west);
\draw[flow] (refactor.east) -- node[edge,above=3pt] {signature contract} (judge.west);
\draw[feedback] (judge.south) -- (10.60,-5.30)
  -- node[edge,above=2pt] {revise} (3.40,-5.30) -- (refactor.south);
\draw[flow] (judge.east) -- (14.10,-4.65) -- (14.10,-2.70)
  -- (2.60,-2.70) -- (planner.north);

\node[stage] at (.35,-6.60) {Certify};
\node[review,text width=6.1cm,minimum height=.85cm] (auditors) at (4.10,-7.40)
  {Source / semantic auditor\\Adversarial / causal auditor};
\node[review,text width=2.8cm] (arbiter) at (11.70,-7.40) {Arbiter};
\node[role,text width=5.6cm] (goal) at (3.65,-8.95)
  {Goal refiner $\leftrightarrow$ Goal judge};
\node[role,text width=5.4cm] (certrepair) at (10.60,-8.95)
  {Refactor $\leftrightarrow$ Loop judge};
\draw[flow,preaction={draw=white,line width=2.2pt}]
  (prover.east) -- (9.00,-3.20) -- (9.00,-2.50)
  -- (14.75,-2.50) -- (14.75,-6.00)
  -- node[edge] {closed candidate} (4.10,-6.00) -- (auditors.north);
\draw[flow] (auditors.east) -- (arbiter.west);
\draw[feedback] (arbiter.south) -- (11.70,-8.18)
  -- node[edge] {revision required} (3.65,-8.18) -- (goal.north);
\draw[flow] (goal.east) -- (certrepair.west);
\draw[feedback] (certrepair.south) -- (10.60,-9.40)
  -- (.35,-9.40) -- (.35,-7.40) -- (auditors.west);

\node[stage] at (.35,-10.34) {Learn};
\node[role,text width=5.6cm] (extract) at (3.65,-11.00)
  {Extraction planner $\rightarrow$ Extractor};
\node[review,text width=5.6cm] (libjudge) at (3.65,-12.15)
  {Library judge $\leftrightarrow$ Rewriter};
\node[review,text width=5.6cm] (remediate) at (3.65,-13.35)
  {Remediator $\leftrightarrow$ Fidelity judge};
\node[memory,text width=4.8cm] (knowledge) at (10.60,-11.00)
  {Knowledge synthesizer\\$\updownarrow$\\Knowledge judge};
\node[role,text width=4.8cm,minimum height=.85cm] (merge) at (10.60,-13.50)
  {Merge\\library and caller builds};
\draw[flow] (arbiter.east) -- (14.10,-7.40) -- (14.10,-9.84)
  -- node[edge] {certified development and history} (3.65,-9.84)
  -- (extract.north);
\draw[flow] (10.60,-9.84) -- (knowledge.north);
\draw[flow] (extract.south) -- (libjudge.north);
\draw[feedback] (libjudge.south) -- node[edge,right=2pt] {caller revision} (remediate.north);
\draw[feedback] (remediate.west) -- (.35,-13.35) -- (.35,-12.15) -- (libjudge.west);
\draw[flow] (libjudge.east) -- (7.40,-12.15) -- (7.40,-13.50) -- (merge.west);
\draw[flow] (knowledge.south) -- node[edge,right=2pt] {approved records} (merge.north);

\node[memory,text width=10cm,minimum height=.75cm] (library) at (7.25,-15.20)
  {SOptLib: Lean declarations and construction records};
\draw[flow] (merge.south) -- (10.60,-14.70) -| (library.north);
\draw[feedback] (library.east) -- (15.25,-15.20) -- (15.25,1.90)
  -- (2.60,1.90) -- (model.north);
\node[edge,rotate=90] at (15.25,-10.0) {retrieval for subsequent tasks};
\node[align=center,text=black!65,
  font=\fontfamily{lmss}\selectfont\fontsize{8}{10}\selectfont,
  text width=14cm] at (7.25,-16.10)
  {Shared tools: Lean inspection and checking, symbol search, LeanSearch.\\
   Solid arrows: handoffs. Dashed arrows: review feedback or reuse.};
\end{tikzpicture}
\caption{\textbf{Agent roles and review loops.}
Roles are grouped by pipeline stage. Model combines object-model refactoring with
a modeling audit and an alignment judge. The Reconstruct / Refactor panel contains
the Planner--Prover--Audit proof loop and the blocker-triggered Refactor--Judge
contract loop. Certify combines independent semantic audits with adjudication and
a localized revision loop; Learn reviews reusable declarations before integration. The upstream Source producer supplies the source package to Model.
Arrows summarize handoffs; execution follows the applicable review verdicts and
build checks.}
\label{fig:agent-topology}
\end{figure}

The prompt instructions require source evidence and affected consumers for each
reconstruction, and distinguish stated assumptions from facts requiring proof.
Retrieval targets a coherent mathematical step, described by its desired relation
and decisive hypotheses. Complete prompt builders and output
schemas accompany the engine.

\subsubsection{Operational Judge discipline}
\label{app:judge-discipline}

The Judge prompt is an admission policy over a proposed reconstruction. We report
its structure here rather than reproducing the full operational text; the complete
prompt and output schema are included in the accompanying artifact package.

\begin{table}[!htbp]
\caption{Structure of the Phase~2b Judge prompt.}
\label{tab:judge-prompt-structure}
\centering
\small
\begin{tabular}{@{}>{\RaggedRight\arraybackslash}p{.20\linewidth}>{\RaggedRight\arraybackslash}p{.72\linewidth}@{}}
\toprule
\indexlabel{Input} & Source text, current Lean state, blocker, prior reviews, and compilation status. \\
\indexlabel{Decision} & Return either \texttt{matches\_paper} or \texttt{needs\_more\_refactor}. \\
\indexlabel{Faithfulness} & Preserve the source-facing theorem head; classify every new premise and record affected downstream consumers. \\
\indexlabel{Proof handoff} & Require a proved surrogate, a compiler-grounded voucher, or a source-grounded literature debt for each remaining root cause. \\
\indexlabel{Failure rule} & Fail closed when the structural cause, source evidence, or next proof route is unresolved. \\
\bottomrule
\end{tabular}
\end{table}

The prompt supplies the semantic policy; source-facing contracts, signature
diffs, fresh verification, and endpoint checks provide the mechanical containment.


\subsection{Worked Example: SAM's Gaussian KL Bridge}
\label{app:sam-construction}

This worked example follows a local obligation in SAM's generalization proof
through diagnosis, reconstruction, and proof completion. The full argument combines
PAC-Bayes, a Gaussian KL calculation, and a radius bound that connects Gaussian-averaged
empirical loss to neighborhood sharpness. We focus on the KL calculation required
by the PAC-Bayes step.

\paragraph{Target obligation.}
For posterior $Q=\mathcal N(w,\sigma^2 I_d)$ and prior
$P_j=\mathcal N(0,v_j I_d)$, with $\sigma>0$ and $v_j>0$, the proof needs
\[
\operatorname{KL}(Q\Vert P_j)
=\frac12\left(
\frac{d\sigma^2+\lVert w\rVert^2}{v_j}
-d+d\log\frac{v_j}{\sigma^2}\right).
\]
The available Gaussian KL theorem uses Euclidean coordinates. The SAM development
represents the laws as affine pushforwards of a standard Gaussian on a
finite-dimensional inner-product space $E$. Applying the theorem therefore requires
identifying these probability measures across the two representations.

\paragraph{Blocker and routing decision.}
A direct transport attempt required \lean{BorelSpace E} and\newline
\lean{SecondCountableTopology E}, while the original bridge allowed an arbitrary
measurable structure on $E$. That interface did not establish that the orthonormal
coordinate map was a measurable equivalence. Planner classified the obstruction
as an object-model gap and requested reconstruction. The source places parameters
in $W\subseteq\mathbb R^d$; the revised interface makes this Euclidean measurable
structure explicit on the Gaussian bridge and its consumers.

\paragraph{Constructed theory.}
The reconstruction builds the measurable equivalence supplied by an orthonormal basis.
It first proves that an affine standard-Gaussian pushforward equals the corresponding
multivariate Gaussian by matching their means and covariance forms. The lemma
\lean{coordinateGaussianLaw\_map} then identifies the posterior and prior under
the coordinate map. A second lemma, \lean{coordinateGaussianKLTransport}, applies
KL invariance under this measurable equivalence. The source-facing calculation
uses that transport, the available isotropic Gaussian KL formula, and preservation
of the squared norm to prove the displayed identity.

\paragraph{Review and downstream use.}
Judge accepted the canonical Euclidean realization and the proved transport
bridges, checked that the source-facing KL predicates were preserved, and verified
successful compilation with no active proof holes in the SAM file.
The selected-Gaussian prior-grid proof consumes the resulting KL supplier at its
chosen positive scale, without adding the KL identity as a premise. This closes
the local calculation needed by the probability-event construction; the radius
bound and the remaining event comparisons are separate parts of the target proof.

\FloatBarrier

\section{SOptLib Construction, Reuse, and Ablation}
\label{app:soptlib-experiments}

SOptLib combines Lean declarations with a mathematical index. The following
describes their organization and use in the supplied engine, followed by the
evaluation protocol and ablation.

\subsection{Library Organization and Index}
\label{app:soptlib-organization}

The core layout follows the mathematical roles in Table~\ref{tab:layers}:
{\normalsize
\begin{Verbatim}[breaklines=true,breakanywhere=true,breaksymbolleft={}]
SOptLib/
  Model/       Glue/       Layer0/       Layer1/
docs/knowledge/
  CATALOG.md               INDEX.md
\end{Verbatim}
}
The module directory describes mathematical scope and dependencies; the catalog
maps declarations to files, layers, and keywords. The index maps each reusable
declaration to its mathematical concept, source provenance, applicability, and
downstream uses. Code and documentation therefore expose the reusable result and
the conditions under which it can be applied.

\subsection{Retrieval and Use during Formalization}
\label{app:soptlib-retrieval}

\paragraph{Query formation and tool roles.}
Agents formulate queries from the current proof state and the mathematical step
being attempted, then inspect relevant declarations through the mathematical index. Symbol search covers declaration names, signatures, and documentation in
the target, project imports, SOptLib, algorithm examples, and local Mathlib.
Results distinguish visible declarations, candidates requiring an import, and
reference examples. LeanSearch complements this with natural-language queries
for Mathlib results, which agents check against the pinned local environment.
The two tool wrappers are \path{lean_search_symbols.py} and
\path{lean_leansearch.py}.

\paragraph{Heuristic ranking.}
Project and SOptLib declarations use a weighted lexical score after identifier
splitting and stop-word filtering. Each query token contributes 10 for an exact
name-token match or 6 for a prefix match, plus 4 for a signature match, 3 for a
docstring match, and 2 for a source-snippet match. Shared consecutive token pairs
add 3 each; a full-query substring match adds 8. Cross-scope duplicates prefer
target-visible declarations, then imported project declarations, SOptLib,
reference examples, and Mathlib. The merged list is ordered by score, with scope
priority breaking ties. The Mathlib branch uses LeanExplore's returned rank,
converted to $k-r+1$ for rank $r$ among up to $k$ results; the separate LeanSearch
tool returns its provider's distances. These scores prioritize inspection;
applicability is established by signature comparison and Lean checking.

\paragraph{Semantic search annotation format.}
Each promoted SOptLib declaration carries a compact, human-readable index next to
its Lean name and signature. The index is designed for discovery: the declaration
and its type remain Lean's authority, while the colored fields expose the concept,
proof role, provenance, and intended reuse conditions.

\begin{table}[!htbp]
\caption{Fields in the natural-language index attached to an SOptLib declaration.}
\label{tab:soptlib-index-fields}
\centering
\small
\begin{tabular}{@{}>{\RaggedRight\arraybackslash}p{.21\linewidth}>{\RaggedRight\arraybackslash}p{.70\linewidth}@{}}
\toprule
\indexlabel{Concept} & Mathematical statement and informal meaning. \\
\indexlabel{Layer / gap} & Library layer and the proof obligation addressed. \\
\indexlabel{Proof idea} & Short route description and decisive APIs. \\
\indexlabel{Source} & Mathlib, SOptLib, or source-theorem provenance. \\
\indexlabel{Used in} & Downstream proof steps and algorithm families. \\
\indexlabel{Origin} & Algorithm that motivated the reusable entry and, when available, its search note. \\
\bottomrule
\end{tabular}
\end{table}

\begin{center}
\setlength{\fboxsep}{6pt}
\fcolorbox{rulegray}{panelbg}{%
\begin{minipage}{0.91\linewidth}
\small
\indexlabel{Concept}\quad Pointwise gradients with a Lipschitz gradient give a smooth
quadratic upper bound on a convex feasible segment.\par
\indexlabel{Layer / gap}\quad Layer0; Level 1, smooth-objective descent.\par
\indexlabel{Proof idea}\quad Restrict to the affine segment, apply the
\texttt{HasGradientAt} chain rule, and integrate the Lipschitz derivative bound.\par
\indexlabel{Source}\quad Convex-segment, chain-rule, and Hilbert-space
Cauchy--Schwarz APIs.\par
\indexlabel{Used in}\quad Stochastic conditional-gradient and mirror-descent
steps before oracle-error or prox-descent absorption.\par
\indexlabel{Declaration}\par
\path{smooth_quadratic_upper_bound_of_hasGradientAt_lipschitzOn_convex}
\end{minipage}}
\end{center}

The box is a discovery index, not a separate Lean source listing. Full comments,
signatures, and proof terms remain in the accompanying Lean files; the paper shows
only the fields needed to understand how an agent finds and checks a declaration.

\paragraph{Prompt guidance for reuse.}
Before decomposing a proof into local algebra or regularity calculations,
agents search for the largest coherent reusable step supported by the available
premises. The query retains the desired relation and decisive conditions while
abstracting away paper-specific names and incidental encodings. Agents inspect
nearby constructors and companion lemmas for promising hits. A partial match
leads to a refined query for the missing relation or a smaller reusable component.
The prompt prioritizes preservation of the source objects, assumptions, and
conclusion when choosing an interface.

\paragraph{From a candidate to a proof step.}
Agents inspect the complete signature, defining module, and neighboring APIs
before reuse. A matching result is instantiated; differences in coordinates,
indexing, or representation require a proved bridge. Names, imports, and the
resulting application are checked in Lean. The accompanying structured-search
utilities can retain query scopes, candidate signatures, and import-check
outcomes for review. Live searches follow the evolving proof state.

\subsection{Extraction, Review, and Integration}
\label{app:soptlib-integration}

Extraction generalizes reusable objects and arguments from certified developments,
proves them in a staging area, and checks their instantiation in the originating
algorithm. Review examines source correspondence, generality, and overlap with
existing APIs. Merge integrates accepted declarations, updates their algorithm-
level uses, and rebuilds the library, originating development, and promoted
examples. The declaration catalog is regenerated after the checks pass.

\subsection{Evaluation Protocol}

Each FOML task retrieves SOptLib support accumulated before its start.
Accepted declarations from a completed development become available to later
FOML tasks through the integration process above.

All five research-paper tasks use the same fixed SOptLib version, containing only
definitions and lemmas accumulated from the FOML developments.

\paragraph{Comparison scope.}

The main comparison evaluates each system in its specified configuration,
including its tools and available mathematical support. The 43-case Judge
and Planner--Audit ablation examines diagnosis and repair at selected
obstructions, testing faithfulness control. The SOptLib ablation examines
final artifact quality and construction resources.

\subsection{Library Census}
\label{app:library-census}

The SOptLib release contains 42 Lean files and \NSOptLOC{} lines, including
its 44-line top-level entry file. Its \NDecls{} source declaration heads
include private declarations and exclude automatically generated constants. Of
these, \NLemmas{} are theorems. The catalog contains \NLibCatalog{} index
entries. Across the sources selected for release,
imported local theory and staging support contribute 6,857 physical lines in
48 distinct file contents, deduplicated by hash and counted separately from
algorithm-local code and SOptLib.

\subsection{Formalization without SOptLib}

Each paired comparison starts by preparing the same task materials. The Full
configuration then retains SOptLib, while the Removed configuration isolates it
for the subsequent formalization and proof construction with Lean and Mathlib.
Both configurations use the same source material, target statements, and
Planner--Audit--Judge workflow. The paired tasks are AMSGrad, NSAGD, SBMD,
and NSMD; each has human and GPT ratings and token-usage measurements.

\paragraph{Quality.}
Human scores are blind ratings of the final artifacts on the 1--7 rubric. GPT
columns report the 0--100 G-Eval and FidelityEval scores from the main
evaluation. Human ratings are identical between Full and Removed for the four
completed pairs. The AMSGrad and NSAGD Removed GPT entries average three
assessments per protocol.

\par\noindent\begin{minipage}{\textwidth}
\captionof{table}{
Full retains SOptLib and Removed isolates it after the paired materials are
prepared. Scores use the same human and GPT evaluation protocols as the main
evaluation. AMSGrad and NSAGD Removed GPT scores average three assessments per
protocol. Each $\Delta$ is Removed minus Full. The mean averages rows with both
Full and Removed scores.}
\label{tab:soptlib-ablation-results}
\centering
\setlength{\tabcolsep}{3pt}
\renewcommand{\arraystretch}{1.14}
\begin{tabularx}{\textwidth}{@{}lRRRRRRRRR@{}}
\toprule
& \multicolumn{3}{c}{Human (1--7)}
& \multicolumn{3}{c}{GPT G-Eval}
& \multicolumn{3}{c}{GPT FidelityEval} \\
\cmidrule(lr){2-4}\cmidrule(lr){5-7}\cmidrule(l){8-10}
Task & Full & Removed & $\Delta$ & Full & Removed & $\Delta$ & Full & Removed & $\Delta$ \\
\midrule
AMSGrad & 7 & 7 & 0.0 & 95.3 & 94.7 & -0.6 & 85.3 & 89.7 & +4.4 \\
SBMD & 7 & 7 & 0.0 & 90.7 & 90.0 & -0.7 & 93.3 & 92.7 & -0.6 \\
NSMD & 6 & 6 & 0.0 & 94.0 & 93.0 & -1.0 & 89.3 & 88.7 & -0.6 \\
NSAGD & 7 & 7 & 0.0 & 95.7 & 97.3 & +1.6 & 84.3 & 83.7 & -0.6 \\
\midrule
Mean & 6.8 & 6.8 & 0.0 & 93.9 & 93.8 & -0.2 & 88.1 & 88.7 & +0.7 \\
\bottomrule
\end{tabularx}

\end{minipage}\par

\paragraph{Token usage.}
Table~\ref{tab:soptlib-token-usage} reports token usage for the paired
formalization and proof-construction runs.

\par\noindent\begin{minipage}{\textwidth}
\captionof{table}{Token usage for the paired formalization and proof-construction
runs, in millions (M). Total is input plus output; cached input is included in
Total. Output includes reasoning tokens.}
\label{tab:soptlib-token-usage}
\centering
\setlength{\tabcolsep}{4pt}
\begin{tabularx}{\textwidth}{@{}lRRRRRR@{}}
\toprule
& \multicolumn{2}{c}{Total tokens (M)}
& \multicolumn{2}{c}{Cached input (M)}
& \multicolumn{2}{c}{Output tokens (M)} \\
\cmidrule(lr){2-3}\cmidrule(lr){4-5}\cmidrule(l){6-7}
Task & Full & Removed & Full & Removed & Full & Removed \\
\midrule
AMSGrad & 177.90 & 191.67 & 166.11 & 178.80 & 1.087 & 1.217 \\
SBMD & 156.981 & 309.956 & 144.915 & 278.413 & 1.091 & 2.606 \\
NSAGD & 129.45 & 195.81 & 94.76 & 181.85 & 0.753 & 1.447 \\
NSMD & 364.921 & 926.769 & 332.026 & 869.105 & 2.914 & 5.004 \\
\bottomrule
\end{tabularx}

\end{minipage}\par

Counts cover planning, proving, and internal review in the paired runs.

\paragraph{Interpretation of the pattern.}
Table~\ref{tab:soptlib-ablation-results} shows similar final ratings across
configurations, while the Removed runs use more tokens: 7.7\% for AMSGrad,
97.5\% for SBMD, 51.3\% for NSAGD, and 154.0\% for NSMD.
SOptLib contributes through two complementary channels. Its mathematical
index exposes source boundaries and applicability conditions for route planning
and candidate adjudication. Its verified Lean declarations and bridges provide
reusable domain facilities,
so later agents can instantiate established infrastructure rather than rebuild
the same mathematical connections.

\end{document}